\documentclass{article}

\usepackage{microtype}
\usepackage{graphicx}
\usepackage{comment}
\usepackage{multirow,multicol}
\usepackage{subcaption}
\usepackage{ulem}
\usepackage{makecell}
\usepackage{booktabs} 

\usepackage[accepted]{icml2025}

\usepackage{amsmath}
\usepackage{amssymb}
\usepackage{mathtools}
\usepackage{amsthm}

\usepackage[capitalize,noabbrev]{cleveref}

\theoremstyle{plain}

\theoremstyle{definition}

\theoremstyle{remark}

\definecolor{ForestGreen}{HTML}{3e9c2a}
\usepackage[textsize=tiny]{todonotes}

\icmltitlerunning{\textbf{\textcolor{ForestGreen}{CLOVER}}: \textcolor{ForestGreen}{C}ross-\textcolor{ForestGreen}{L}ayer \textcolor{ForestGreen}{O}rthogonal \textcolor{ForestGreen}{Ve}cto\textcolor{ForestGreen}{r}s}

\begin{document}

\twocolumn[
\icmltitle{
           \textbf{\textcolor{ForestGreen}{CLOVER}}: \textcolor{ForestGreen}{C}ross-\textcolor{ForestGreen}{L}ayer \textcolor{ForestGreen}{O}rthogonal \textcolor{ForestGreen}{Ve}cto\textcolor{ForestGreen}{r}s Pruning and Fine-Tuning
           }



\begin{icmlauthorlist}
\icmlauthor{Fanxu Meng}{sch,comp}
\icmlauthor{Pingzhi Tang}{sch}
\icmlauthor{Fan Jiang}{sch}
\icmlauthor{Muhan Zhang}{sch,comp}
\end{icmlauthorlist}

\icmlaffiliation{sch}{Institute for Artificial Intelligence, Peking University}
\icmlaffiliation{comp}{State Key Laboratory of General Artificial Intelligence, Peking University}
\icmlkeywords{Machine Learning, ICML}

\vskip 0.3in
]




\begin{abstract}
Decoder-only models generate tokens autoregressively by caching key/value vectors, but as the cache grows, inference becomes memory-bound. To address this issue, we introduce CLOVER (Cross-Layer Orthogonal Vectors), a novel approach that treats pairs of attention layers as a set of low-rank decompositions. CLOVER applies Singular Value Decomposition (SVD) to the \( Q \)-\( K \) and \( V \)-\( O \) pairs within each attention head. The resulting singular values can either guide pruning or serve as trainable parameters for efficient fine-tuning of all orthogonal vectors. After pruning or fine-tuning, these values are reintegrated into the model without increasing its parameter count.
We apply CLOVER to various models, including GPT-2 XL, DeepSeek-V2-Lite, Whisper-Large-v3, Stable Diffusion XL, and LLaMA-3.2-11B-Vision. Our results demonstrate that CLOVER significantly improves pruning efficiency. For instance, the perplexity of pruning 70\% of the \( Q \)-\( K \) pairs in GPT-2 XL is similar to that of pruning just 8\% with vanilla methods. Fine-tuning the singular values further results in a full-rank update, outperforming state-of-the-art methods (LoRA, DoRA, HiRA, and PiSSA) by 7.6\%, 5.5\%, 3.8\%, and 0.7\%, respectively, on eight commonsense tasks for LLaMA-2 7B.
\end{abstract}

\begin{figure}[!t]
    \centering
    \begin{subfigure}[b]{0.495\columnwidth}
        \includegraphics[width=\columnwidth]{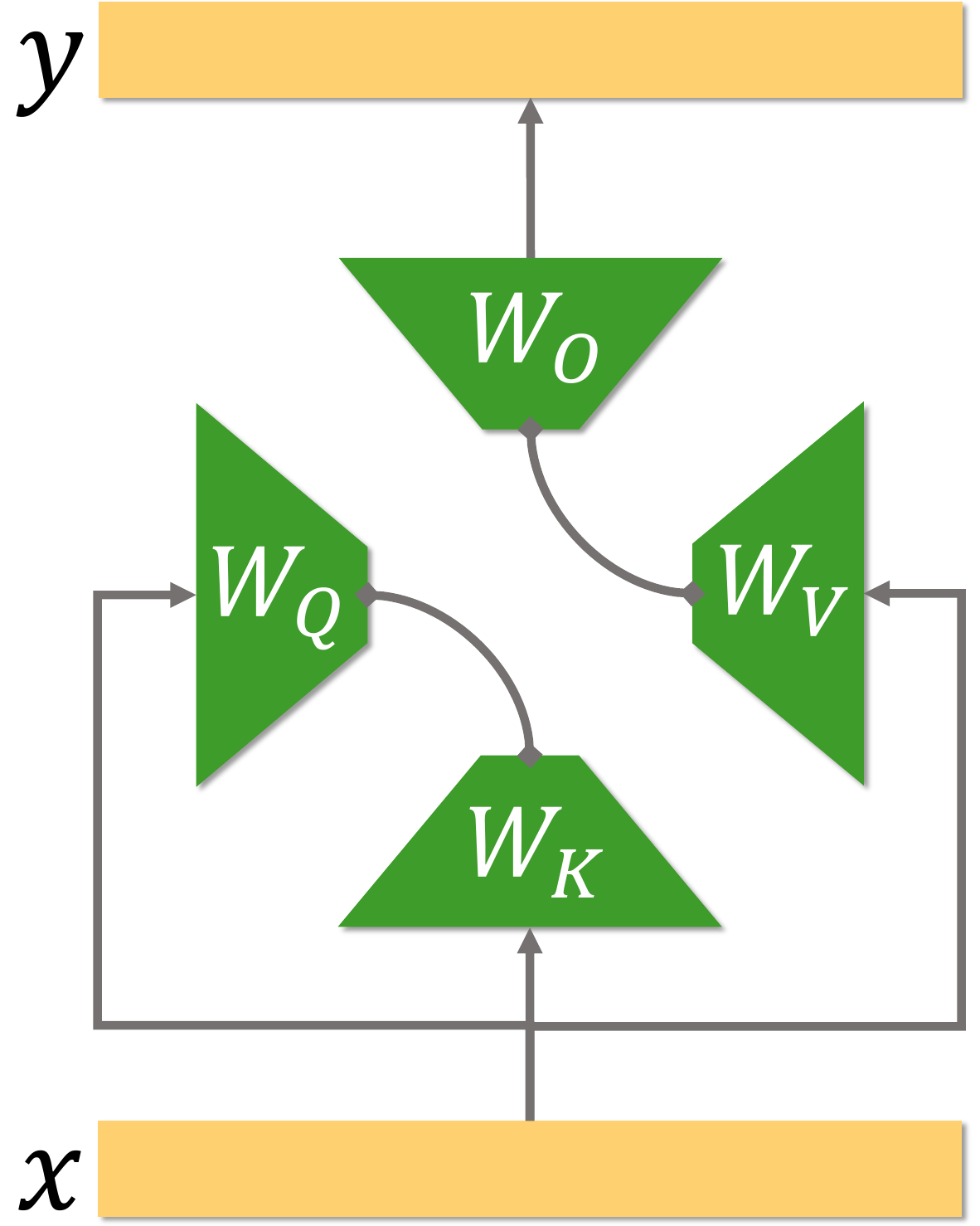}
        \caption{Multi-Head Attention}
        \label{subfig:fullft}
    \end{subfigure}
    \hfill
    \begin{subfigure}[b]{0.495\columnwidth}
        \includegraphics[width=\columnwidth]{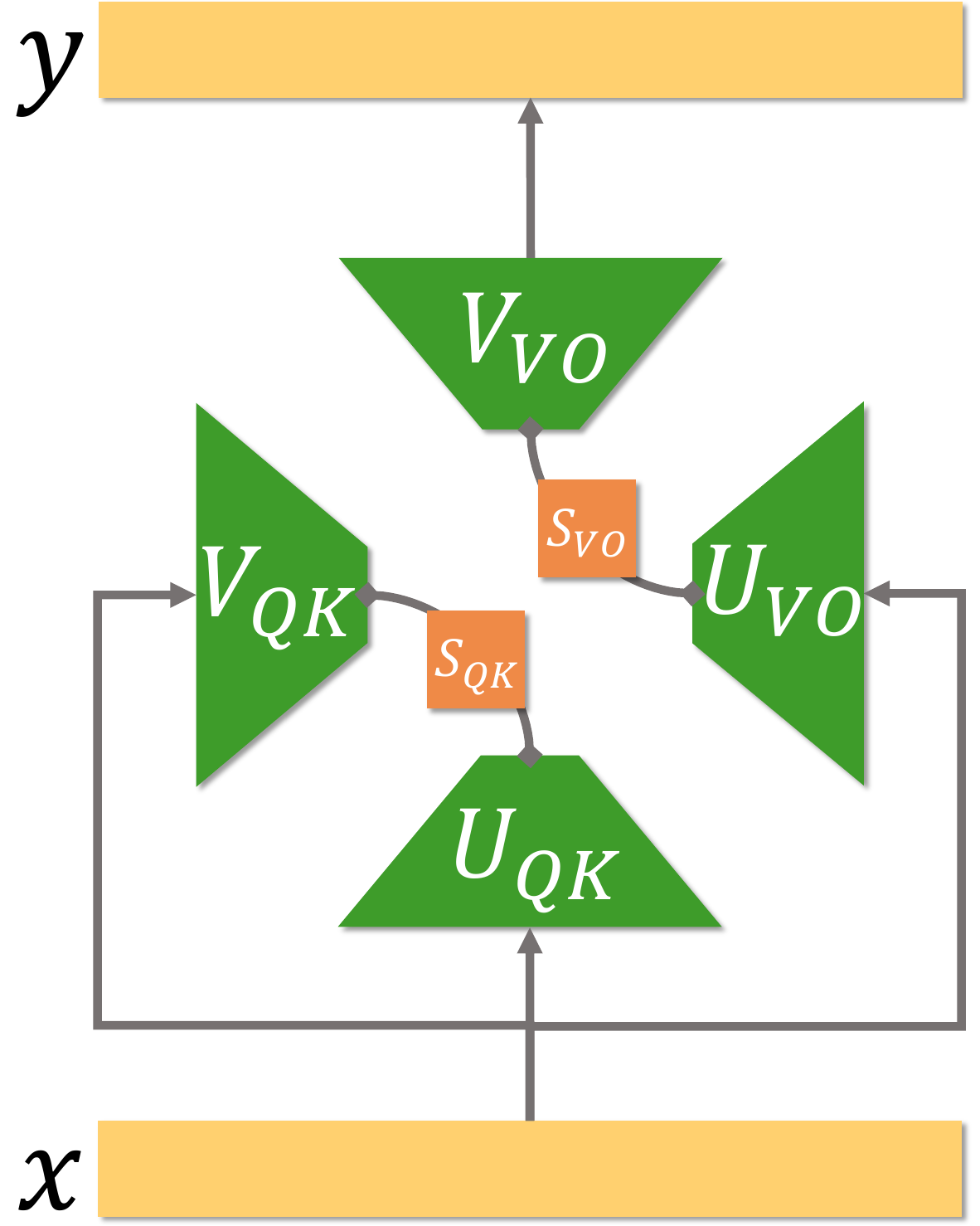}
        \caption{CLOVER}
        \label{subfig:clover}
    \end{subfigure}
        \begin{subfigure}[b]{0.495\columnwidth}
        \includegraphics[width=\columnwidth]{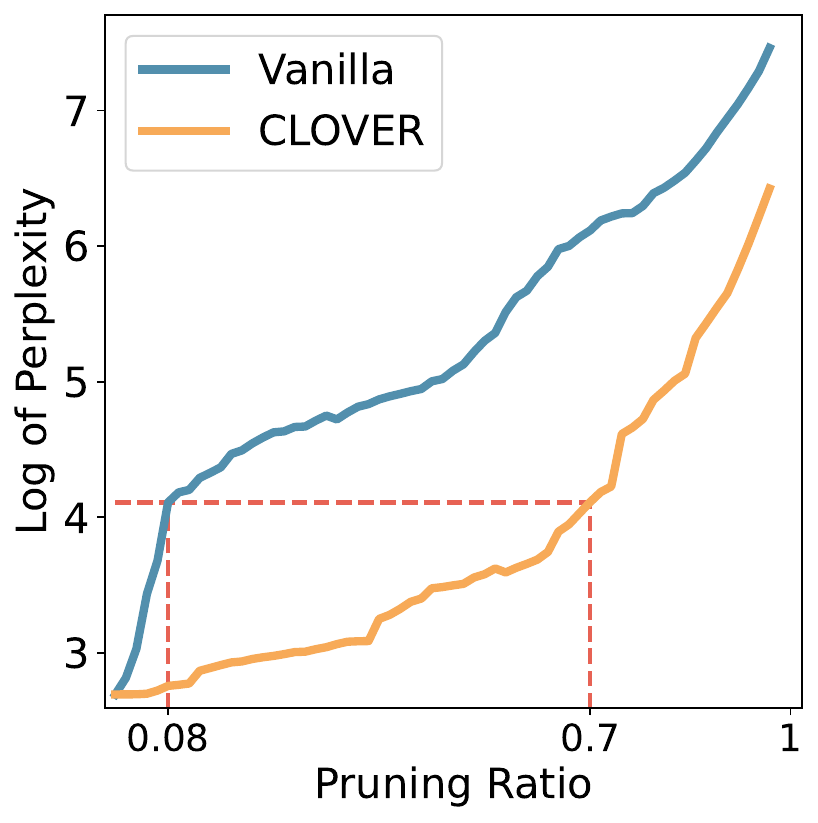}
        \caption{Pruning without Training}
        \label{subfig:qk_ppl}
    \end{subfigure}
    \hfill
    \begin{subfigure}[b]{0.495\columnwidth}
        \includegraphics[width=\columnwidth]{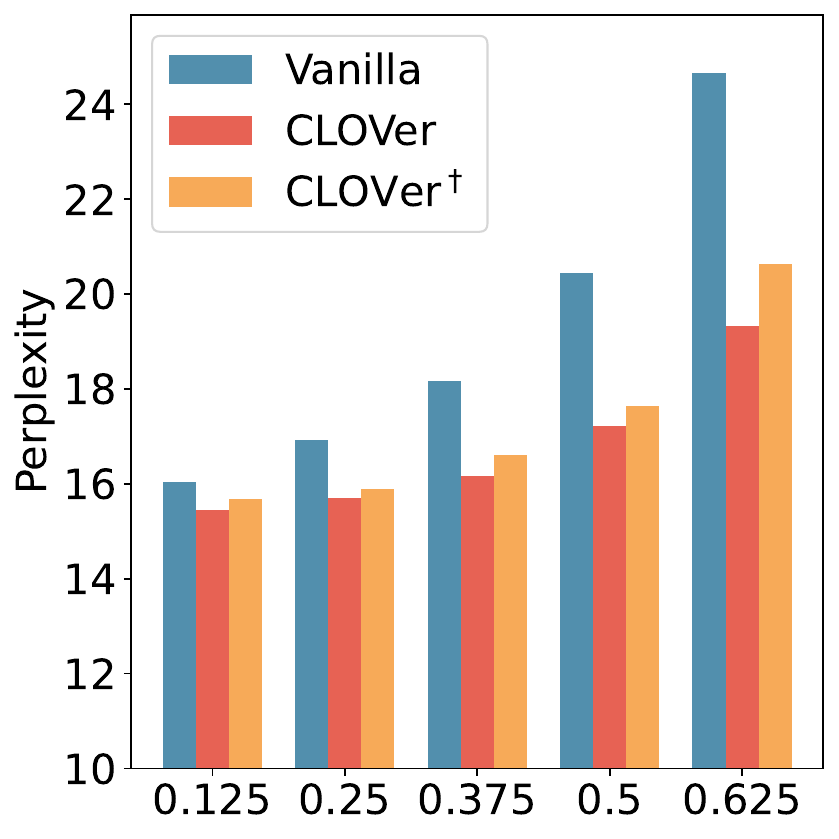}
        \caption{Fine-Tuning Pruned Model}
        \label{subfig:finetune_qk_ppl}
    \end{subfigure}
    \caption{(a) We treat the Query-Key and Value-Output layers within a single attention head as a unified structure. (b) Apply SVD to obtain two sets of singular vectors for initializing the Q-K and V-O layers, along with singular values that guide pruning or enable efficient full-rank fine-tuning. (c) This cross-layer orthogonalization strategy allows for higher pruning rates. (d) The pruned model maintains strong performance after fine-tuning.}
    \label{fig:clover}
\end{figure}

\section{Introduction}
In recent years, Large Language Models (LLMs) have rapidly evolved into essential tools for productivity \cite{gpt4o, claude35sonnet, team2024gemini}. Open-source models \cite{llama3, mixtral8x22b, qwen2_5, dsviii, team2024gemma, abdin2024phi} have also narrowed the performance gap with closed-source models. The success of LLMs is largely attributed to Next Token Prediction \cite{radford2018improving, brown2020language}, where tokens are predicted sequentially, with attention computed between each token and all preceding ones. To avoid redundant computations, key-value features are cached. However, as model size grows, the overhead of caching becomes substantial, leading to memory and communication bottlenecks. For instance, a 65B parameter model \cite{llama} with 8-bit key-value quantization requires over 86GB of GPU memory to store 512K tokens, exceeding the capacity of a single H100-80GB GPU \cite{sun2024you}.

To enable efficient training and inference, we introduce CLOVER (Cross-Layer Orthogonal Vectors), a novel method that orthogonalizes the Query, Key, Value, and Output vectors without generating additional transformation matrices. As shown in Figure \ref{subfig:fullft}, we treat the \( Q \)-\( K \) and \( V \)-\( O \) pairs in each attention head as a low-rank decomposition of \( W_{QK} \) and \( W_{VO} \). By crossing these layers and performing SVD on \( W_{QK} \) and \( W_{VO} \), the Query, Key, Value, and Output vectors become orthogonal within each attention head. 
Figure \ref{subfig:clover} illustrates how the resulting singular values can guide pruning or serve as trainable parameters for efficient fine-tuning. After pruning or fine-tuning, these values can be reintegrated into the model without increasing its parameter count. Notably, previous methods, such as SVFT~\cite{lingam2024svft}, obtain orthogonal vectors by directly performing orthogonal decomposition on the matrix at each layer, which results in an accompanying transformation matrix, doubling the parameter count. In contrast, CLOVER treats the \( Q \)-\( K \) pairs as transformation matrices for each other, and similarly for the \( V \)-\( O \) pairs. CLOVER only generates a small set of singular values to guide pruning and fine-tuning, which can be merged back into the model without increasing inference costs.

\textbf{By orthogonalizing the vectors, we eliminate linear redundancy.} Attention heads contain numerous non-zero norm vectors. Directly pruning these vectors would degrade performance, but orthogonalizing them allows us to represent the entire attention head's space using a small set of orthogonal bases. The remaining vectors are nearly zero, making them safe to prune. As shown in Figure \ref{subfig:qk_ppl}, pruning an average of 45 vectors in the query-key pair using CLOVER results in a perplexity similar to that of vanilla pruning, which prunes only 5 vectors.
Moreover, CLOVER generates a singular value matrix between the \( Q \)-\( K \) and \( V \)-\( O \) pairs. By updating this matrix during fine-tuning, \textbf{CLOVER learns linear combinations of all orthogonal bases within each attention head.} In contrast, PiSSA can only learn from a subset of orthogonal vectors, potentially causing some data projections to approach zero in those directions, leading to non-functional adapters during training. As shown in Figure \ref{subfig:finetune_qk_ppl}, fine-tuning a very small number of singular values can achieve performance close to that of fine-tuning all attention heads.
We summarize the contribution of our paper as follows:

\begin{itemize}
    \item We treat the Q-K and V-O pairs in each attention head as low-rank approximations of \(W_{QK} \) and \(W_{VO} \). By performing SVD, we orthogonalize the attention head without adding extra transformation matrices.
    \item This orthogonalization reduces linear redundancy, is compatible with any pruning method, and allows for higher pruning ratios. Pruning 46.42\% of the vectors in Whisper’s attention head preserves performance without requiring additional training.
    \item CLOVER enables efficient full-rank updates, surpassing SOTA methods such as LoRA, DoRA, HiRA, and PiSSA on eight commonsense reasoning tasks across LLaMA 7B/13B, LLaMA-2-7B, and LLaMA-3-8B, with additional analyses highlighting its advantages.
\end{itemize}

\section{Related Work}
\paragraph{LLM Compression}
To mitigate the high memory demands of KV Caches in long-context models, several techniques have been proposed. These include reducing sequence length with linear attention \cite{katharopoulos2020transformers, wang2020linformer, peng2023rwkv, gu2023mamba, de2024griffin}, dynamic token pruning \cite{fu2024lazyllm, jo2024a2sf, li2024snapkv}, compressing the key-value rank \cite{shazeer2019fast, ainslie2023gqa, liu2024deepseek, yu2024effectively}, and pruning head dimensions \cite{ashkboos2024slicegpt, xia2023sheared, sun2023simple}. Additional approaches include sharing key-value representations across layers \cite{sun2024you, brandon2024reducing, liu2024minicache, zuhri2024mlkv} and quantizing KV cache weights and activations \cite{frantar2022gptq, dettmers2022gpt3, xiao2023smoothquant, liu2024kivi, hooper2024kvquant}.
Among them, structure pruning is hardware-friendly but can reduce performance when non-zero dimensions are removed \cite{ma2023llm}. Fine-tuning can recover some of the lost performance, but it's computationally expensive. To address this, Parameter Efficient Fine-Tuning (PEFT) methods are used \cite{guo2023compresso}.

\paragraph{Parameter Efficient Fine-Tuning.}
Several strategies have been introduced to minimize fine-tuning parameters while maintaining performance. These include low-rank adaptation \cite{hu2021lora}, partial-parameter fine-tuning \cite{zaken2021bitfit, lawton2023neural, zhao2020masking, sung2021training, ansell2021composable, xu2021raise, guo2020parameter, fu2023effectiveness}, soft prompt fine-tuning \cite{hambardzumyan2021warp, lester2021power, li2021prefix, liu2023gpt, vu2021spot, asai2022attempt, wang2023multitask}, and sparse matrix fine-tuning \cite{qiu2023controlling, liu2023parameter, yuan2024bridging}.
Among these, LoRA is widely used due to its simplicity and effectiveness, with recent works enhancing it further \cite{zhang2023adalora, zi2023delta, liu2024dora, zhao2024galore, jiang2024mora}. PiSSA \cite{meng2024pissa} improves convergence speed by initializing adapters with principal singular values and vectors, also reducing quantization error \cite{wang2024loraga, wang2024lorapro, li2024svdquant}. However, PiSSA is limited by its use of a fixed set of orthogonal bases.
SVFT \cite{lingam2024svft} directly applies Singular Value Decomposition (SVD) to the original matrix, but this increases the number of parameters, raising computational overhead and reducing efficiency.
The CLOVER method addresses these issues by treating the Query-Key pairs in each attention head as low-rank matrices. Using orthogonal decomposition, CLOVER eliminates the need for additional transformation matrices. Instead, it leverages a small set of singular values to linearly combine orthogonal vectors, making the approach more parameter-efficient. After fine-tuning, the adapter can be smoothly reintegrated into the original matrix structure.

\section{\textbf{\textcolor{ForestGreen}{CLOVER}}: \textcolor{ForestGreen}{C}ross-\textcolor{ForestGreen}{L}ayer \textcolor{ForestGreen}{O}rthogonal \textcolor{ForestGreen}{Ve}cto\textcolor{ForestGreen}{r}s}
\label{sec:clover_method}

Below is a step-by-step explanation of CLOVER method and explain why it can update orthogonal decompose the Query, Key, Value, Output layers in Multi-Head Attention without need introduce any transfer matrix. We mainly use the computation of the $Q$-$K$ pair in as an example. Then extended to the $V$-$O$ pair.

\paragraph{Multi-Head Self-Attention Setup.}
In a multi-head self-attention mechanism with \(H\) heads, each head \(h \in \{1, \dots, H\}\) computes an attention score as:

\[
\text{attn}(Q_h, K_h) 
= \text{softmax}\!\Bigl(\tfrac{Q_h K_h^\top}{\sqrt{d}}\Bigr),
\]

where 
\(H\) is the number of attention heads,
\(d\) is the dimensionality of each head,
\(X \in \mathbb{R}^{n \times D}\) is the input matrix (\(n\) is the sequence length, \(D\) is the total hidden dimension),
\(Q_h, K_h \in \mathbb{R}^{n \times d}\) are the query and key representations for head \(h\),
\(W_Q, W_K \in \mathbb{R}^{D \times H \times d}\) are weights for projecting the input \(X\) into queries and keys.

Specifically, the queries and keys for head \(h\) are obtained by multiplying \(X\) with the corresponding “slice” of \(W_Q\) and \(W_K\), respectively:

\[
Q_h = X \, W_Q^{[:, h, :]}, 
\quad 
K_h = X \, W_K^{[:, h, :]}.
\]

\paragraph{Cross Layers Merging.}

Substituting \(Q_h\) and \(K_h\) into \(Q_h K_h^\top\), we have: 
\[
Q_h K_h^\top = X \, W_Q^{[:, h, :]} \,\bigl(W_K^{[:, h, :]}\bigr)^\top X^\top.
\]

Notice that the original weights \(W_Q^{[:, h, :]}\) and \(W_K^{[:, h, :]}\) are each in \(\mathbb{R}^{D \times d}\), once multiplied together, the resulting matrix \(W_{QK}^h \;=\; W_Q^{[:, h, :]} \,\bigl(W_K^{[:, h, :]}\bigr)^\top \) has dimension \(D \times D\).
Since \(d \ll D\), using \(W_{QK}^h\) directly in computations—or storing it as trainable parameters—would be highly inefficient, limiting the use cases of such parameter merging.

\paragraph{Cross Layers Orthogonal Decomposition}

To address the large size of \(W_{QK}^h\), we factorize \(W_{QK}^h\) via SVD:

\[
W_{QK}^h 
\;=\; U_{QK}^h \; S_{QK}^h \; V_{QK}^h,
\]

where
\(U_{QK}^h\) is a \(D \times D\) orthogonal matrix,
\(S_{QK}^h\) is a \(D \times D\) diagonal matrix of singular values,
\(V_{QK}^h\) is another \(D \times D\) orthogonal matrix.

Since \(W_Q^{[:, h, :]}\) and \(W_K^{[:, h, :]}\) each have shape \(\mathbb{R}^{D \times d}\), the rank of \(W_{QK}^h\) is at most \(d\). Thus the actual \textbf{non-zero singular values} in \(S_{QK}^h\) are \textbf{at most} \(d\). We can truncate the SVD to keep only the top-\(r\) singular values without loss:
\[
W_{QK}^h 
\;=\;
U_{QK}^h [:, :r] \; S_{QK}^h [:r, :r] \; \bigl(V_{QK}^h [:, :r]\bigr)^\top,
\]
where \(r \le d\).

The process can be easily applied to $W_V$ and $W_O$, as introduced in Appendix \ref{sec:vo_ortho}.
\paragraph{CLOVER for Pruning}

After performing SVD, we can rewrite the weight matrix \( W_{QK}^h \) as follows:

\[
W_{QK}^h \;=\; 
\underbrace{U_{QK}^h [:, :r] \, S_{QK}^h [:r, :r]}_{\tilde{U}^h \in \mathbb{R}^{D \times r}} 
\;
\underbrace{\bigl(V_{QK}^h [:, :r]\bigr)^\top}_{\tilde{V}^h \in \mathbb{R}^{r \times D}}.
\]

Instead of storing the full matrices \( W_Q^h \) and \( W_K^h \in \mathbb{R}^{D \times d} \), we store the smaller factors \( \tilde{U}^h \in \mathbb{R}^{D \times r} \) and \( \tilde{V}^h \in \mathbb{R}^{r \times D} \), which can be significantly smaller than the original matrix since \( r \leq d \ll D \). This leads to a reduction in memory usage and computational cost. Additionally, we can \textbf{prune} singular values (and their corresponding singular vectors) below a chosen threshold. This further reduces the parameter count and computational overhead.

\paragraph{CLOVER for Fine-Tuning}

CLOVER can be used not only for pruning, but also for parameter-efficient fine-tuning. We freeze the matrices \( U_{QK}^h [:, :r] \) and \( V_{QK}^h [:, :r] \), and only fine-tune the singular values \(S_{QK}^h [:r, :r]\).

In contrast to SVFT, which factorizes the entire weight matrices \( W_Q, W_K, W_V, W_O \in \mathbb{R}^{D \times D} \) individually, CLOVER factorizes the merged weights $W_{QK}^h$ and $W_{OV}^h$ within each attention head, significantly reducing the parameters.
By applying SVD factorization within each attention head, CLOVER constrains the effective rank of the cross-layer matrix to \( d \). As a result, the tunable matrix \( S_{QK} \) has a size bounded by \( \mathbb{R}^{H \times d \times d} \) (considering all heads). In comparison, SVFT requires factorizing large matrices each into three components (\( U, S, V \in \mathbb{R}^{D \times D}\)), leading to a significant increase in parameter count and computational overhead, even with sparse updates for the singular values \( S \).

For example, consider the LLaMA 2-7B model with \( H = 32 \) attention heads and a head dimension of \( d = 128 \). By factorizing each head separately, the largest size for \( S_{QK} \) is \( \mathcal{O}(32 \times 128 \times 128) \), which is significantly smaller than factorizing a \( \mathbb{R}^{4096 \times 4096} \) matrix. This makes CLOVER’s parameter efficiency comparable to that of a LoRA configuration with rank 32, as shown in Appendix \ref{sec:Hyperparameters}, but with additional potential for pruning.

\begin{table*}[ht]
\caption{Pruning GPT-2-XL's attention layers with CLOVER and vanilla pruning at various ratios, evaluating perplexity on Wikitext2 (lower is better), and fine-tuning on OpenWebText with different token budgets. The base model's perplexity is 14.78.}

\vskip 0.1in
\setlength{\tabcolsep}{1.2mm}
\centering
\begin{tabular}{c|cc|ccc|ccc}
\toprule
\multirow{2}{*}{Pruning Ratio}&\multicolumn{2}{c|}{\textbf{w/o Training Perplexity($\downarrow$)}}&\multicolumn{3}{c|}{\textbf{66M Tokens Perplexity ($\downarrow$)}}&\multicolumn{3}{c}{\textbf{131M Tokens Perplexity  ($\downarrow$)}}\\
&Vanilla	&CLOVER	&Vanilla	&CLOVER	&CLOVER$^\dagger$ &Vanilla	&CLOVER	&CLOVER$^\dagger$\\
\midrule
12.5\% & 33.76 & \textbf{15.89} & 16.04 & \textbf{15.45} & 15.67 & 16.38 & 15.77 & \textbf{15.42} \\
25.0\% & 78.36 & \textbf{17.45} & 16.93 & \textbf{15.70} & 15.89 & 17.07 & 16.05 & \textbf{15.75} \\
37.5\% & 159.4 & \textbf{20.95} & 18.17 & \textbf{16.17} & 16.60 & 18.14 & 16.48 & \textbf{16.41} \\
50.0\% & 338.9 & \textbf{35.12} & 20.45 & \textbf{17.22} & 17.63 & 19.02 & \textbf{17.13} & 17.71 \\
62.5\% & 538.5 & \textbf{85.25} & 24.65 & \textbf{19.32} & 20.64 & 21.44 & \textbf{18.40} & 20.39 \\
75.0\% & 708.8 & \textbf{187.4} & 36.04 & \textbf{24.65} & 29.28 & 27.22 & \textbf{20.99} & 28.44 \\
\bottomrule
\end{tabular}
\label{tab:clover_sota_pruning}
\vskip -0.1in
\end{table*}

\section{Experiments}
As detailed in Section \ref{sec:clover_method}, CLOVER is highly effective for both pruning and fine-tuning. We presents a series of experiments to validate these capabilities.
In Section \ref{subsec:pruning_gpt2xl}, we compare CLOVER with Vanilla pruning on a GPT-2-XL model \cite{radford2019language}. CLOVER results in less performance degradation, while Vanilla pruning significantly harms the model's performance, making recovery difficult even with fine-tuning.
In Section \ref{subsec:clover_for_finetuning_on_commonsense}, we conduct fine-tuning experiments on eight commonsense tasks, comparing CLOVER with state-of-the-art methods. The results show the effectiveness of CLOVER’s linear combinations of all orthogonal vectors.
In Section \ref{subsec:visualize_pruning_models}, CLOVER is applied to various models. We visualize how it removes linear redundancy between vectors, enabling more efficient pruning.
In Section \ref{subsec:pruning_whisper}, we demonstrate CLOVER’s ability to perform significant pruning on the Whisper model, which exhibits substantial linear redundancy, without requiring fine-tuning.
In Section \ref{subsec:full_direction_projection}, we explain the importance of learning from all the orthogonal vectors by analyzing the projection of data features onto different directions in the model.
In Section \ref{subsec:full_rank_update}, we confirm CLOVER’s full-rank update capability by visualizing the singular value distribution of $\Delta W$ from various methods.
Finally, in Section \ref{subsec:intrusive_dimensions}, we show how CLOVER fine-tunes the model using its inherent properties, without introducing ``intrusive dimension'' like LoRA, which may risk model degradation \cite{shuttleworth2024lora}. 
\subsection{CLOVER for Large Ratio Pruning}
\label{subsec:pruning_gpt2xl}
Due to the need to compute attention between each token and all preceding tokens, compressing attention—particularly the key-value layers—is crucial, despite the larger number of parameters in the MLP. CLOVER represents each attention head with a small number of vectors. Since it only modifies the initialization, it can be combined with any other pruning technique. This paper validates the proposed method using basic structured pruning on GPT-2-XL, rather than targeting state-of-the-art performance. We initialize GPT-2-XL with CLOVER, then prune small singular values based on their magnitude. To maintain inference efficiency, we apply the same pruning rate across all layers, removing a fixed percentage of the smallest singular vectors. The singular values, \(S\), are then merged into the \(U\) and \(V\) matrices. For comparison, we also prune without CLOVER orthogonalization, using \(L_2\)-norms for pruning. After pruning, we evaluate perplexity on the WikiText-2 \cite{merity2016pointer} dataset.
We then fine-tune the pruned models on the OpenWebText \cite{Gokaslan2019OpenWeb} dataset following nanoGPT\footnote{https://github.com/karpathy/nanoGPT}. To minimize disruption to the original model, we fine-tune only the pruned attention layers, leaving the MLP, embedding layers, and LM head unchanged. In the CLOVER$^\dagger$ case, after pruning, \(S\) is not immediately merged into the \(U\) and \(V\) matrices but is used for parameter-efficient fine-tuning, with the merging occurring afterward. We adjust the learning rate from 6e-4 to 6e-3 and remove weight decay, while keeping other hyperparameters consistent with the other two methods. 

Based on Table \ref{tab:clover_sota_pruning}, CLOVER causes less damage to the model than Vanilla pruning, as it transfers functionality into fewer orthogonal bases. For example, pruning 50\% of the parameters without further fine-tuning, CLOVER's perplexity only increases by 1.38$\times$, while Vanilla pruning increases by 21.9$\times$. After fine-tuning, CLOVER's performance far exceeds that of Vanilla pruning. Due to its lower model disruption, CLOVER requires fewer tokens for fine-tuning to restore performance (e.g., perplexity with 66M tokens is close to that with 131M tokens), whereas Vanilla pruning needs more tokens, resulting in higher costs and potential degradation in out-of-domain tasks. Furthermore, by fine-tuning only the singular values from the SVD decomposition and the attention layer biases, CLOVER achieves recovery with fewer training resources and parameter changes. At lower pruning rates, CLOVER even outperforms full attention layer training. However, when pruning rates are too high, accuracy loss becomes significant, and the available parameters for fine-tuning become insufficient (e.g., at 75\% pruning, only 0.15\% of the original attention layer parameters are updated).

\begin{table*}[th]
\caption{Accuracy comparison of LLaMA 7B/13B, LLaMA2 7B, and LLaMA3 8B with various PEFT methods on eight commonsense reasoning datasets. Results of LoRA and DoRA are taken from \cite{liu2024dora}. Results of HiRA are taken from \cite{anonymous2025hira}.}
\vskip 0.1in
\setlength{\tabcolsep}{1.2mm}
\centering
\resizebox{0.99\textwidth}{!}{
\begin{tabular}{ccccccccccccc}
\toprule
\textbf{Model} & \textbf{Method} & \textbf{Params} & \textbf{BoolQ} & \textbf{PIQA}&\textbf{SIQA}& \makecell{\textbf{Hella}\\\textbf{Swag}} & \makecell{\textbf{Wino}\\\textbf{Grande}}& \textbf{ARC-e} & \textbf{ARC-c} & \textbf{OBQA} & \textbf{Avg.} \\ \hline
ChatGPT & - & - & 73.1 & 85.4 & 68.5 & 78.5 & 66.1 & 89.8 & 79.9 & 74.8 & 77.0 \\ \hline
\multirow{6}{*}{LLaMA-7B}    
& Series & 0.99\% & 63.0 & 79.2 & 76.3 & 67.9 & 75.7 & 74.5 & 57.1 & 72.4 & 70.8 \\ 
& Parallel & 3.54\% & 67.9 & 76.4 & 78.8 & 69.8 & 78.9 & 73.7 & 57.3 & 75.2 & 72.2 \\ 
& LoRA & 0.83\% & 68.9 & 80.7 & 77.4 & 78.1 & 78.8 & 77.8 & 61.3 & 74.8 & 74.7 \\ 
& DoRA  & 0.84\% & 69.7 &	83.4 &	78.6 &	87.2 &	81.0 &	81.9 &	66.2 &	79.2 &	78.4 \\
& PiSSA  & 0.83\% & \textbf{74.1}& \underline{85.4}& \underline{81.5}& \underline{94.0}& \underline{85.0}& \underline{85.6}& \underline{72.1}& \underline{84.2}& \underline{82.7}\\  
& CLOVER  & 0.83\%& \underline{72.9}&\textbf{86.34}& \textbf{82.1}& \textbf{94.9}& \textbf{85.4}& \textbf{87.5}& \textbf{74.4}& \textbf{86.4}& \textbf{83.7} \\
\hline
\multirow{6}{*}{LLaMA-13B}    
& Series & 0.80\% & 71.8 & 83 & 79.2 & 88.1 & 82.4 & 82.5 & 67.3 & 81.8 & 79.5 \\ 
& Parallel & 2.89\% & 72.5 & 84.9 & 79.8 & 92.1 & 84.7 & 84.2 & 71.2 & 82.4 & 81.4 \\ 
& LoRA & 0.67\% & 72.1 & 83.5 & 80.5 & 90.5 & 83.7 & 82.8 & 68.3 & 82.4 & 80.5 \\ 
& DoRA& 0.68\% & 72.4 & 84.9 & 81.5 & 92.4 & 84.2 & 84.2 & 69.6 & 82.8 & 81.5 \\
& PiSSA  & 0.67\%& \underline{74.6}& \underline{88.0}& \underline{82.9}& \underline{95.5}& \underline{87.0}& \textbf{90.3}& \underline{77.2}& \underline{88.2}& \underline{85.4}\\ 
& CLOVER  & 0.67\%& \textbf{75.2}& \textbf{88.4}& \textbf{83.1}& \textbf{96.0}& \textbf{87.8}& \underline{89.7}& \textbf{79.3}& \textbf{89.8}& \textbf{86.2}\\ 
\hline
\multirow{5}{*}{LLaMA2-7B}    
& LoRA & 0.83\% & 69.8&	79.9&	79.5&	83.6&	82.6&	79.8&	64.7&	81.0&	77.6 \\ 
& DoRA & 0.84\% & 71.8 &83.7&	76.0&	89.1&	82.6&	83.7&	68.2&	82.4&	79.7 \\  
& HiRA & 0.83\% &71.2&	83.4&	79.5& 	88.1&	84.0&	86.7&	  73.8& 	 84.6&	81.4\\
& PiSSA  & 0.83\%& \textbf{75.0}& \textbf{87.0}&\underline{81.6} &\underline{95.0} &\underline{86.5} &\underline{88.5} &\underline{75.9} &\underline{86.4} &\underline{84.5} \\ 
& CLOVER  & 0.83\%& \textbf{75.0}& \underline{86.4}& \textbf{82.0}& \textbf{95.1}& \textbf{87.5}& \textbf{89.6}& \textbf{76.6}& \textbf{89.4}& \textbf{85.2}\\ 
\hline
\multirow{5}{*}{LLaMA3-8B}    
& LoRA & 0.70\% &70.8&	85.2&	79.9&	91.7&	84.3&	84.2&	71.2&	79.0&	80.8 \\ 
& DoRA & 0.71\% & 74.6&	89.3&	79.9&	95.5&	85.6&	90.5&	80.4&	85.8&	85.2 \\ 
& HiRA & 0.70\% &75.4&	\underline{89.7}&	81.2&	95.4&	87.7&	93.3&	\underline{82.9}&	\underline{88.3}&	86.7\\
& PiSSA  & 0.70\%& \textbf{77.2}& \textbf{90.0}& \textbf{82.9}& \underline{96.6}& \underline{88.4}& \textbf{93.6}& 82.4& 87.4& \underline{87.3}\\ 
& CLOVER  & \textbf{0.47\%}& \underline{76.4}& 89.3& \underline{82.1}& \textbf{96.9}& \textbf{89.9}& \textbf{93.6}& \textbf{84.5}& \textbf{90.6}& \textbf{87.9}\\ 
\hline
\end{tabular}}
\label{tab:llama_commonsense}
\vskip -0.1in
\end{table*}
\subsection{CLOVER for Full-Rank Fine-Tuning}
\label{subsec:clover_for_finetuning_on_commonsense}
In this section, we evaluate CLOVER against LoRA \cite{hu2021lora}, DoRA \cite{liu2024dora}, HiRA \cite{anonymous2025hira}, and PiSSA \cite{meng2024pissa} on commonsense reasoning tasks, excluding SVFT \cite{lingam2024svft} due to its significant overhead. The tasks are divided into eight sub-tasks, as outlined in Table \ref{tab:datasets_detail}. Following the DoRA setup, we fine-tune the Commonsense-170k dataset and evaluate each sub-task's test set.
We apply orthogonal decomposition to the Value-Output and fine-tune the resulting singular value matrix. Due to the non-linear RoPE\cite{su2024roformer} operation between the query and key, we perform orthogonal decomposition in the Key layer and fine-tune the transition matrix. Similarly, we treat the 64 consecutive dimensions in the MLP.Up layer as a head, applying orthogonal decomposition and updating the transition matrix. The learnable parameters of LLaMA 7B/13B \cite{llama} and LLaMA-2-7B \cite{llama2} match LoRA/DoRA/HiRA/PiSSA with rank 32 updates. LLaMA-3-8B \cite{llama3} has 2/3 of the trainable parameters compared to the other models.
For a fair comparison, we use the hyperparameters from DoRA (3 epochs, batch size 16, linear scheduler learning rate). We adjusted the learning rate based on DoRA's approach and found that CLOVER performs best with lr=1e-4, which we applied across all models. PiSSA was trained using the same hyperparameters, but with a learning rate of 2e-5, as specified in its original paper. Due to the stable performance of PiSSA and CLOVER during training, we did not perform validation every 80 iterations, as done in DoRA, to select the best-performing model on the validation set for testing. Instead, we trained for the full 3 epochs and used the final model for testing. HiRA’s results are taken directly from its original paper, while the other results are sourced from DoRA's paper.
Table 2 demonstrates that CLOVER consistently outperforms all other methods across all models and tasks. Specifically, on LLaMA 7B, CLOVER outperforms LoRA, DoRA, and PiSSA by 9\%, 5.3\%, and 1\%, respectively. On LLaMA 13B, CLOVER outperforms these methods by 5.7\%, 4.7\%, and 0.8\%. On LLaMA-2-7B, CLOVER surpasses LoRA, DoRA, HiRA, and PiSSA by 7.6\%, 5.5\%, 3.8\%, and 0.7\%. Even on LLaMA-3-8B, with fewer trainable parameters, CLOVER outperforms by 7.1\%, 2.7\%, 1.2\%, and 0.6\%. CLOVER leads in most sub-tasks and ranks second in a few.

\begin{figure*}[!th]
    \centering
    \begin{subfigure}[b]{0.388\columnwidth}
        \includegraphics[width=\columnwidth]{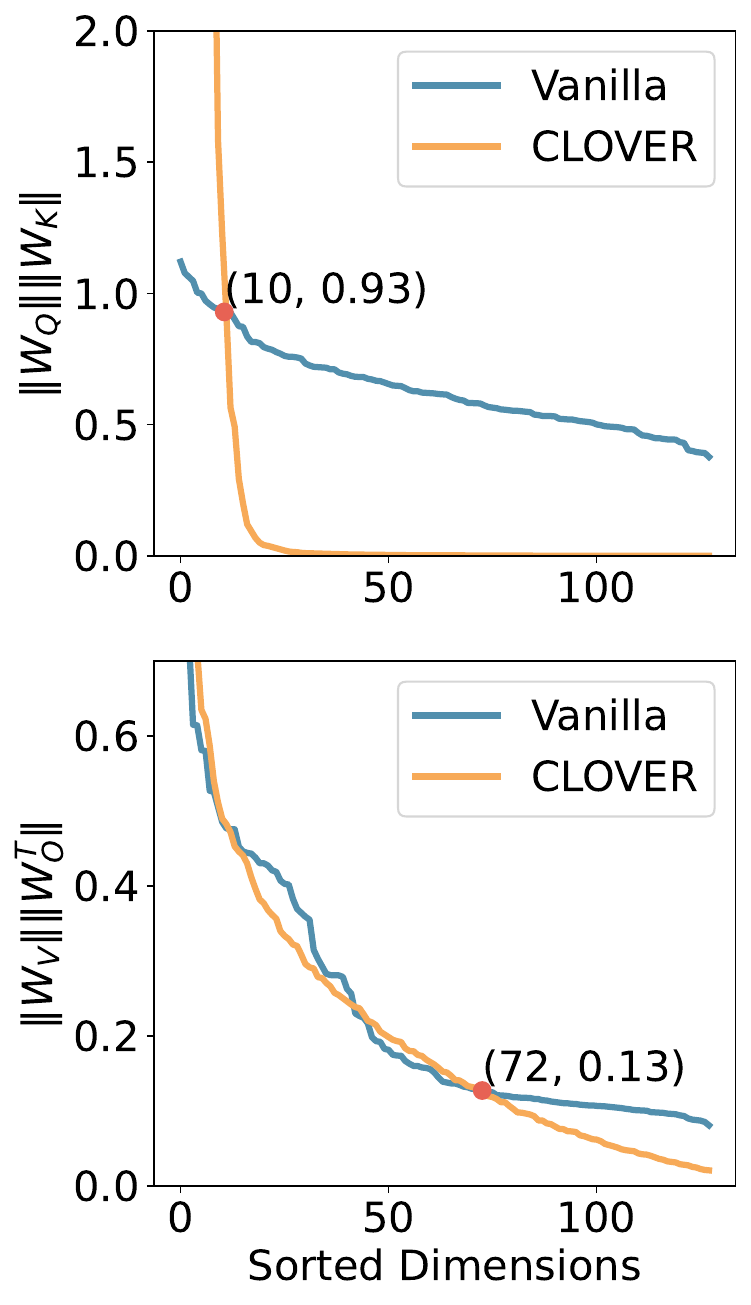}
        \caption{DeepSeek-V2-Lite}
        \label{subfig:deepseek_norm}
    \end{subfigure}
    \hfill
    \begin{subfigure}[b]{0.388\columnwidth}
        \includegraphics[width=\columnwidth]{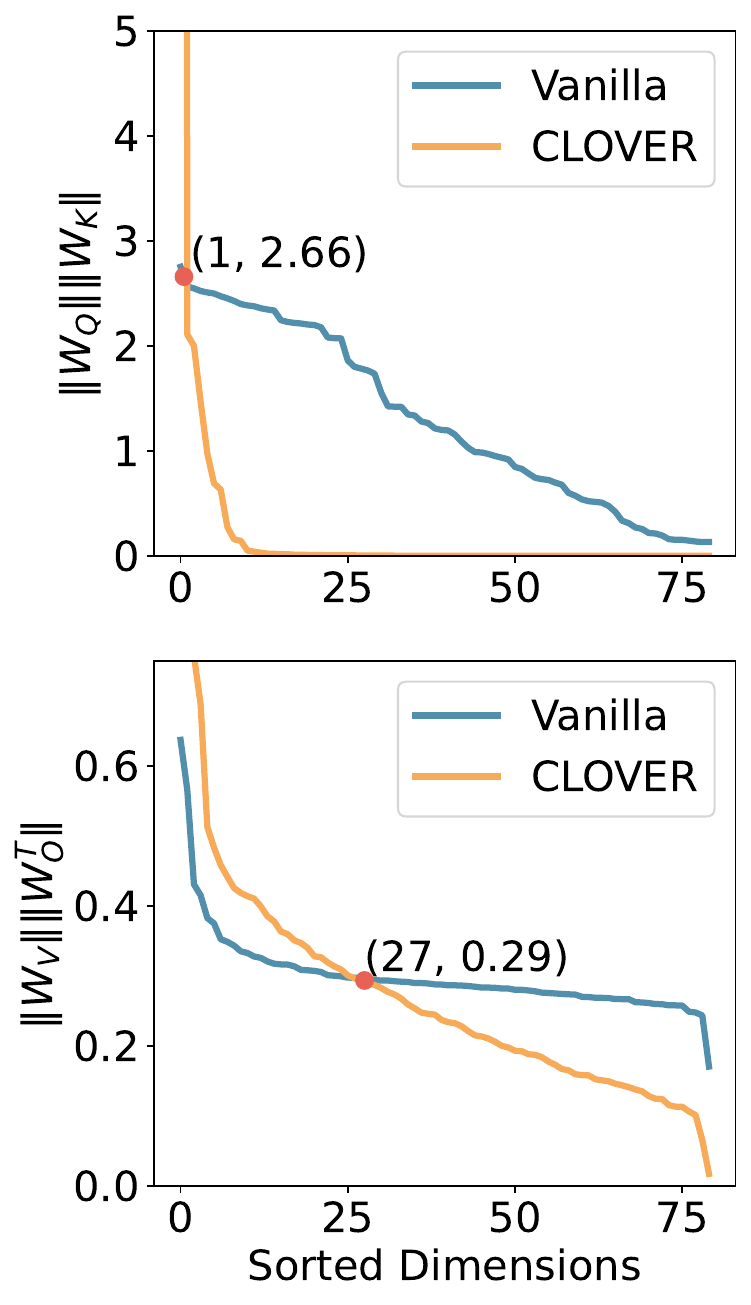}
        \caption{Llama-3.2-Vision}
        \label{subfig:llama3v_norm}
    \end{subfigure}
    \hfill
    \begin{subfigure}[b]{0.4\columnwidth}
        \includegraphics[width=\columnwidth]{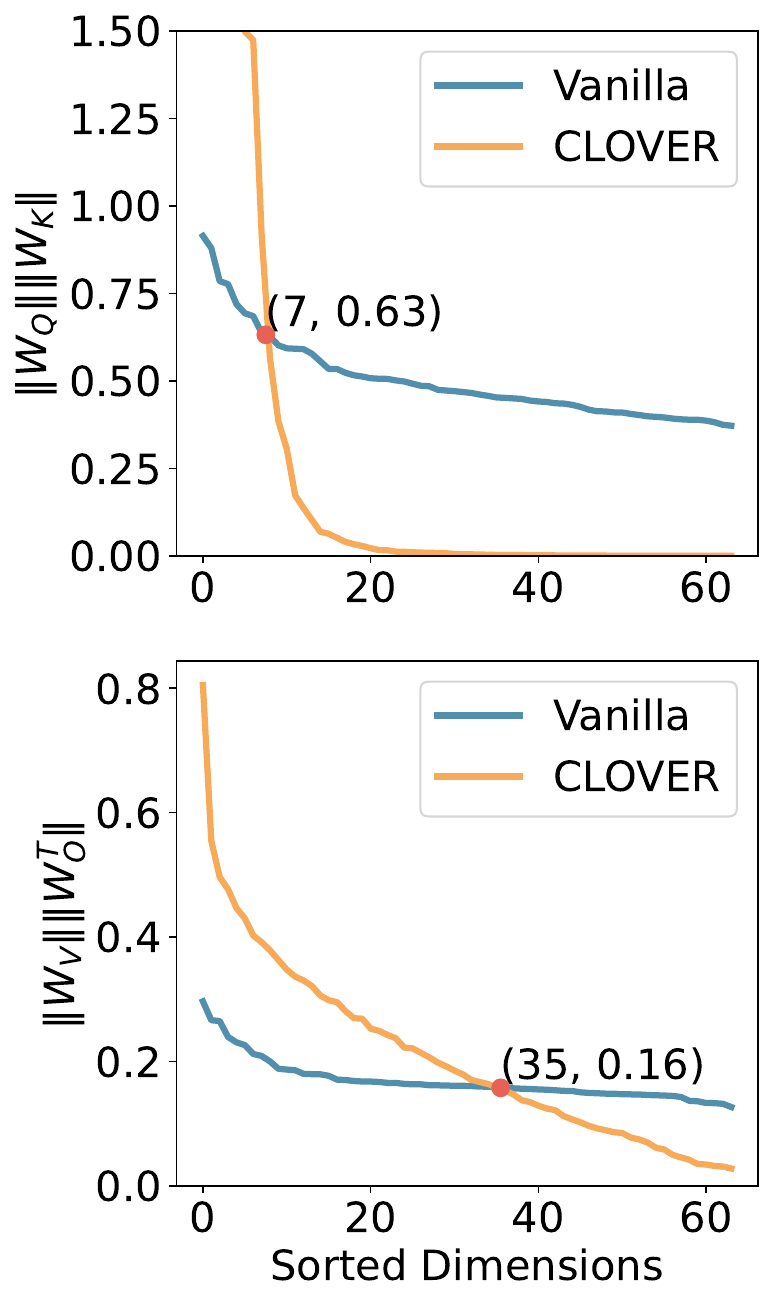}
        \caption{Whisper-Large-v3}
        \label{subfig:wisper_norm}
    \end{subfigure}
    \hfill
    \begin{subfigure}[b]{0.4\columnwidth}
        \includegraphics[width=\columnwidth]{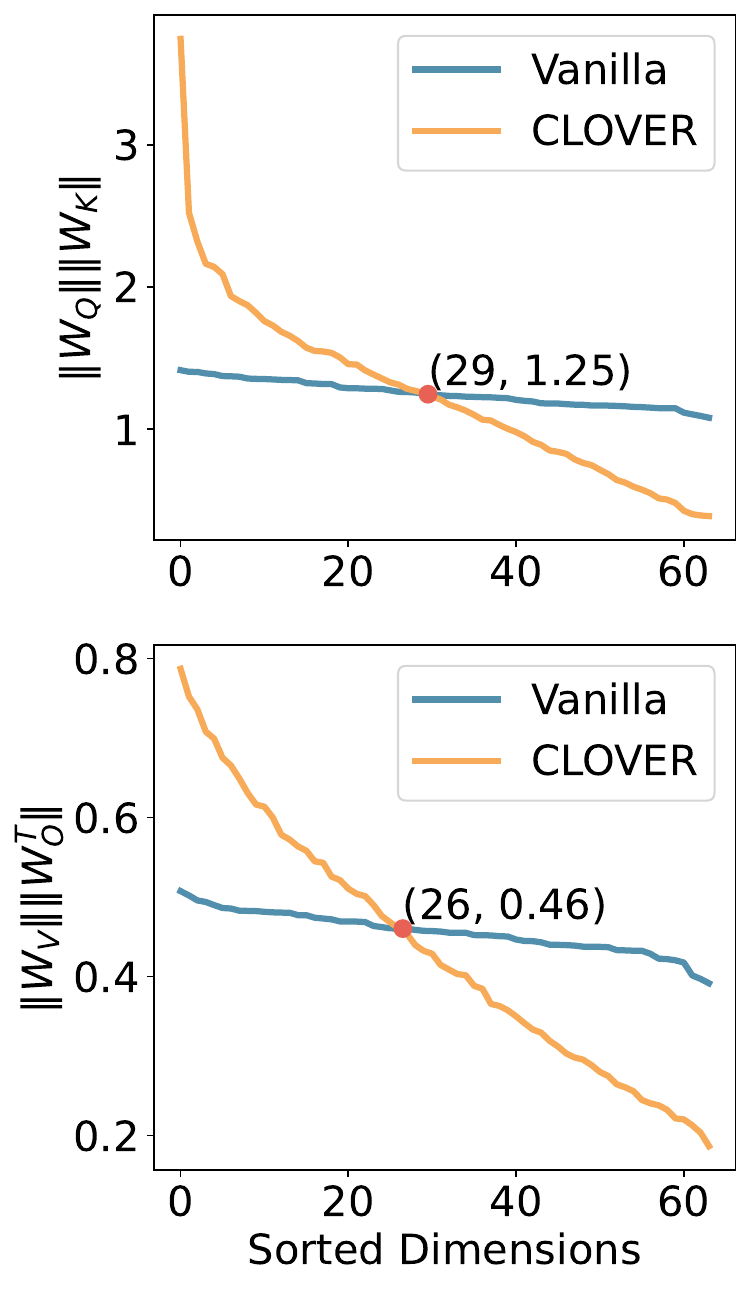}
        \caption{SDXL}
        \label{subfig:gpt2_norm}
    \end{subfigure}
    \hfill
    \begin{subfigure}[b]{0.4\columnwidth}
        \includegraphics[width=\columnwidth]{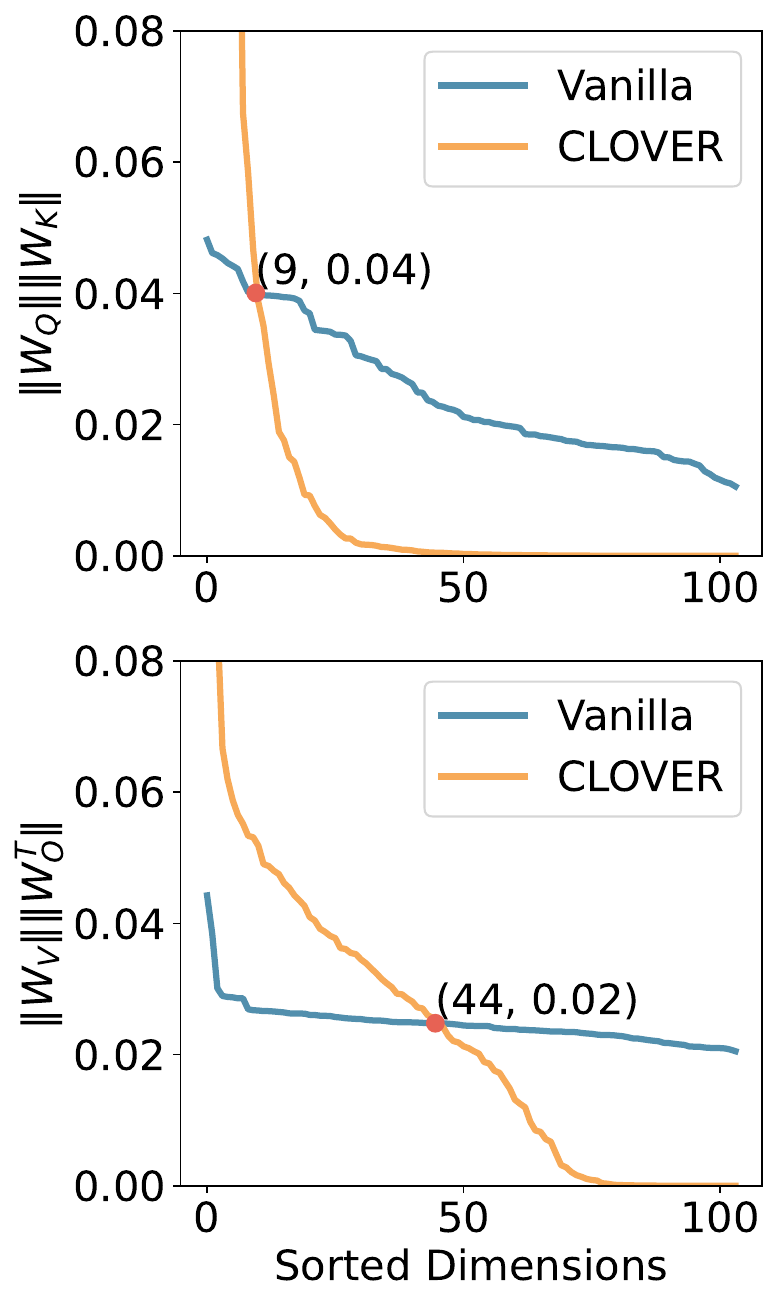}
        \caption{CLIP-ViT-BigG}
        \label{subfig:vit_norm}
    \end{subfigure}
    \caption{CLOVER (orange) uses fewer orthogonal basis vectors than Vanilla Pruning (blue) to span the attention head space. The first row shows the importance of Q-K dimensions, and the second row shows V-O dimensions. After the red dot, CLOVER's importance is lower, and pruning these vectors results in less performance loss.}
    \label{fig:gpt2_wisper_llama3v_norm}
\end{figure*}

\subsection{CLOVER Removal Redundant Vectors}
\label{subsec:visualize_pruning_models}
CLOVER achieves a higher pruning ratio due to the significant linear redundancy present in the model. By representing the entire attention head with only a small number of orthogonal vectors, CLOVER effectively removes this redundancy.
To illustrate the advantages of CLOVER in eliminating linear redundancy, we apply it to a variety range of models, including the large language model DeepSeek-V2-Lite \cite{dsvii}, the multimodal automatic speech recognition and speech translation model Whisper-Large-v3 \cite{radford2023robust}, the multimodal instruction-tuned image reasoning generative models LLaMA-3.2-11B-Vision \cite{llama3}, the image encoder CLIP-ViT-bigG \cite{cherti2022reproducible}, and the image generation model Stable Diffusion XL \cite{podell2023sdxl}.
We compute the $L_2$ norm for each dimension (equal to singular values) in both the Q-K pair and the V-O pair, sorting the values in descending order within each attention head for better visualization. For comparison, we also perform Vanilla Pruning, which does not utilize CLOVER initialization but instead sorts directly based on the $L_2$ norm. 

Figure \ref{fig:gpt2_wisper_llama3v_norm} showcases the first attention head from the first layer of each model.
In the first column of the figure, depicting the Q-K norm, we observe that in the original model, the importance of each dimension is relatively balanced (e.g. Figure \ref{subfig:wisper_norm}). This balanced distribution is a result of the linear redundancy, where different directions are intertwined, making it challenging to prune individual directions without negatively affecting the model's performance.
However, after applying CLOVER’s orthogonal decomposition, only a small number of orthogonal bases on the left side exhibit significantly large norms. These vectors span almost the entire attention head’s space, and the remaining vectors have norms that approach zero, indicating that they are already represented by the dominant singular vectors and can be pruned without loss of performance.
Beyond the red intersection point, CLOVER’s remaining vectors exhibit consistently lower importance than those in Vanilla Pruning, meaning pruning these vectors results in less performance degradation. This demonstrates why CLOVER enables a higher pruning ratio. A similar trend is observed for the V-O pair, although the model’s inherent sparsity is less pronounced than in the Q-K pair, making the effect less noticeable. Still, in most models, pruning half of the vectors has a smaller impact on performance compared to Vanilla Pruning. Notably, in CLIP-ViT-bigG (Figure \ref{subfig:vit_norm}), a proportion of the vectors already have a norm of zero, allowing for safe pruning.

\subsection{CLOVER for Training-Free Pruning}
\label{subsec:pruning_whisper}
As demonstrated by the prominent low-rank properties in Figure \ref{subfig:wisper_norm}, we applied pruning to the Whisper-large-v3 model \cite{radford2023robust}. To intuitively highlight the effectiveness of CLOVER pruning, we present an example using an audio input from the LibriSpeech Long dataset \cite{gandhi2023distil}. For reference, the waveform of this input is shown in Figure \ref{fig:audio}, and the corresponding target translation script is provided in Appendix \ref{target_audio_translate}.

After applying CLOVER to orthogonalize the vectors, we pruned vectors with magnitudes close to zero ($\|W_Q\|\|W_K\| \leq 5 \times 10^{-3}$ and $\|W_V\|\|W_O^\top\| \leq 6 \times 10^{-3}$). This pruning achieved ratios of 56.01\% and 36.82\% for the parameters in $Q$-$K$ Pair and $V$-$O$ Pair, respectively. Remarkably, the model's output remains nearly unchanged, with only one error, which has been highlighted in the text using strikethrough and red for clarity:

\begin{figure}[ht]
    \centering
    \includegraphics[width=\columnwidth]{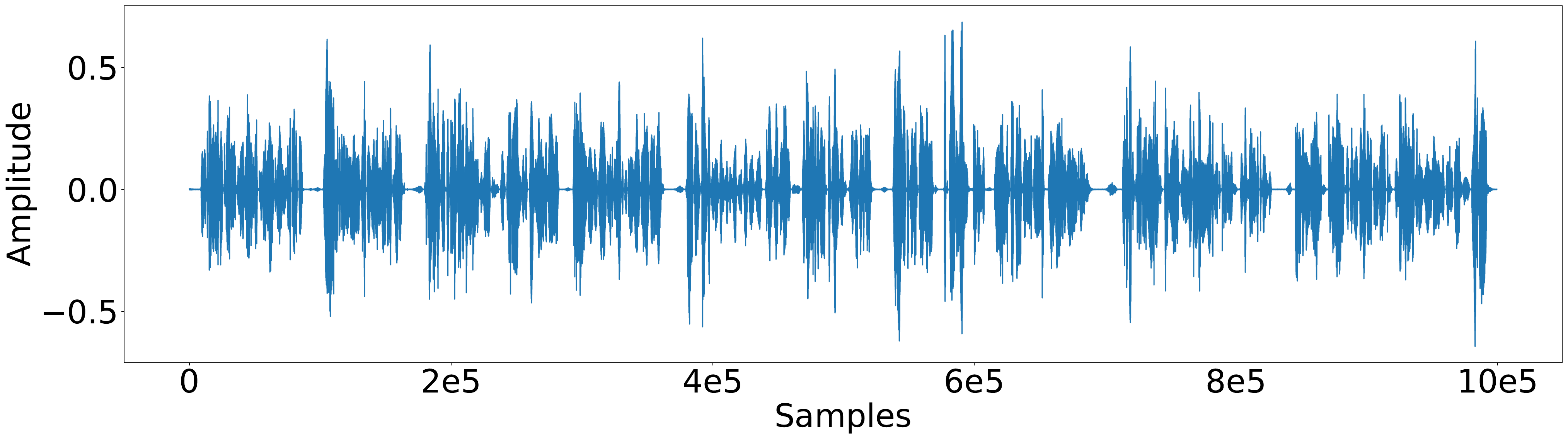}
    \caption{An audio waveform from the librispeech dataset.}
    \label{fig:audio}
\end{figure}

\textit{
Mr. Quilter is the apostle of the middle classes, 
and we are glad to welcome his gospel. 
Nor is Mr. Quilter's manner less interesting than his matter. 
He tells us that at this festive season of the year, 
with Christmas and roast beef looming before us, 
similes drawn from eating and its results occur most readily to the mind. 
He has grave doubts whether Sir Frederick Layton's work is really Greek after all, 
and can discover in it but little of rocky Ithaca. 
Linnell's pictures are a sort of Up Guards and Adam paintings, 
and Mason's exquisite idles are as national as a jingo poem. 
Mr. Birkett Foster's landscapes smile at one much 
in the same way that Mr. Carker used to flash his teeth\sout{. And}\textcolor{red}{, and} Mr. John Collier gives his sitter a cheerful slap on the 
back before he says, like a shampooer in a Turkish bath, next man.
}

In contrast, using a vanilla pruning method with the same pruning ratio, the model completely fails to produce valid outputs:

\textit{
... ... ... ... ... ... ... ... ... ... ... ... ... ... ... ... ... ... ... ... ... ...
}

This example validates our earlier claim that straightforward pruning of non-zero dimensions can lead to accumulated loss. In contrast, CLOVER effectively eliminates linear redundancy, enabling a significantly higher pruning ratio. When the linear redundancy is sufficiently pronounced, CLOVER can even achieve a high pruning ratio without the need for fine-tuning to recover performance.

\subsection{Necessity of Full-Direction Fine-Tuning}
\label{subsec:full_direction_projection}
Besides pruning with a large ratio, CLOVER is capable of learning linear combinations of all orthogonal vectors within each attention head. This capability allows CLOVER to resemble full-parameter fine-tuning more closely.
To highlight the advantages of updating all orthogonal bases, we randomly sampled 16 instances from the Commonsense dataset, fed them into the model, and performed SVD to the model. We then recorded the projection magnitudes of input features along all orthogonal directions. Figure \ref{fig:lora_pissa_direction} visualizes the results for the middle layer, revealing the following insights:
\begin{figure}[!t]
    \centering
    \begin{subfigure}[b]{0.46\columnwidth}
    \includegraphics[width=\columnwidth]{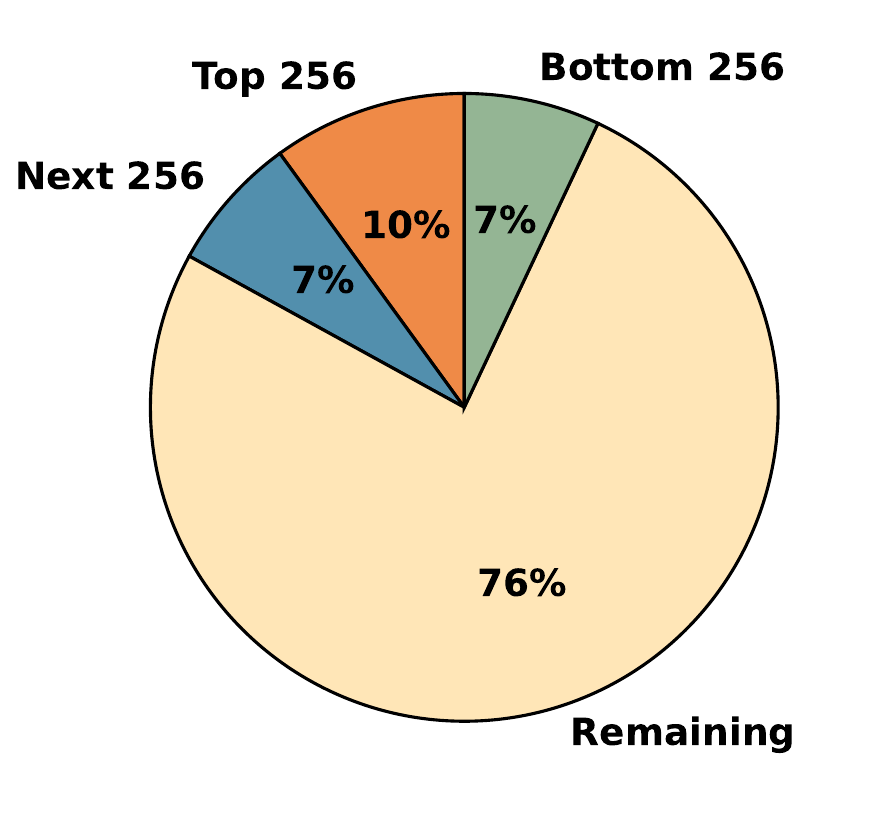}
    \caption{PiSSA}
    \label{subfig:pissa_direction}
    \end{subfigure}
    \begin{subfigure}[b]{0.43\columnwidth}
    \includegraphics[width=\columnwidth]{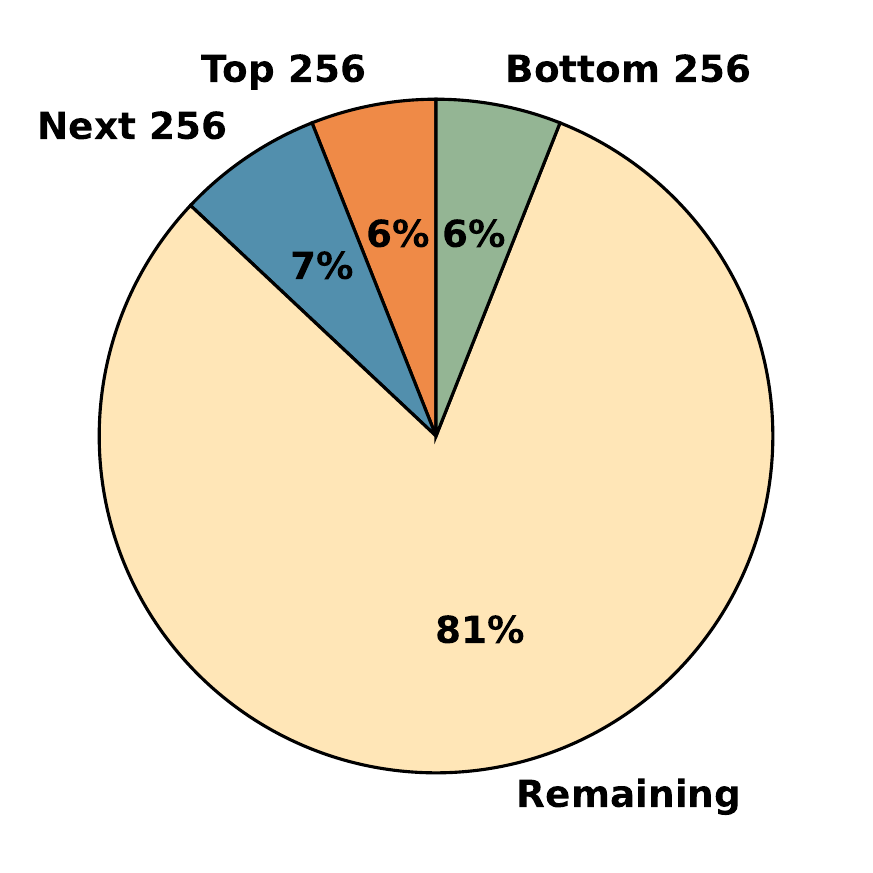}
    \caption{LoRA}
    \label{subfig:lora_direction}
    \end{subfigure}
    \begin{subfigure}[b]{0.5\columnwidth}
    \includegraphics[width=\columnwidth]{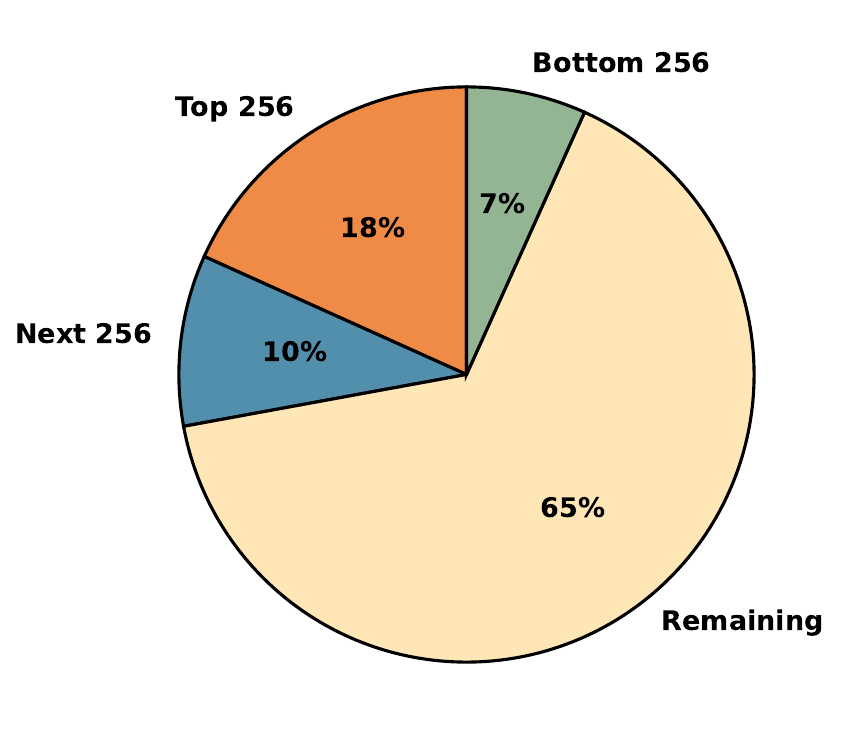}
    \caption{PiSSA with Singular Value}
    \label{subfig:pissa_with_s_direction}
    \end{subfigure}
        \begin{subfigure}[b]{0.44\columnwidth}
    \includegraphics[width=\columnwidth]{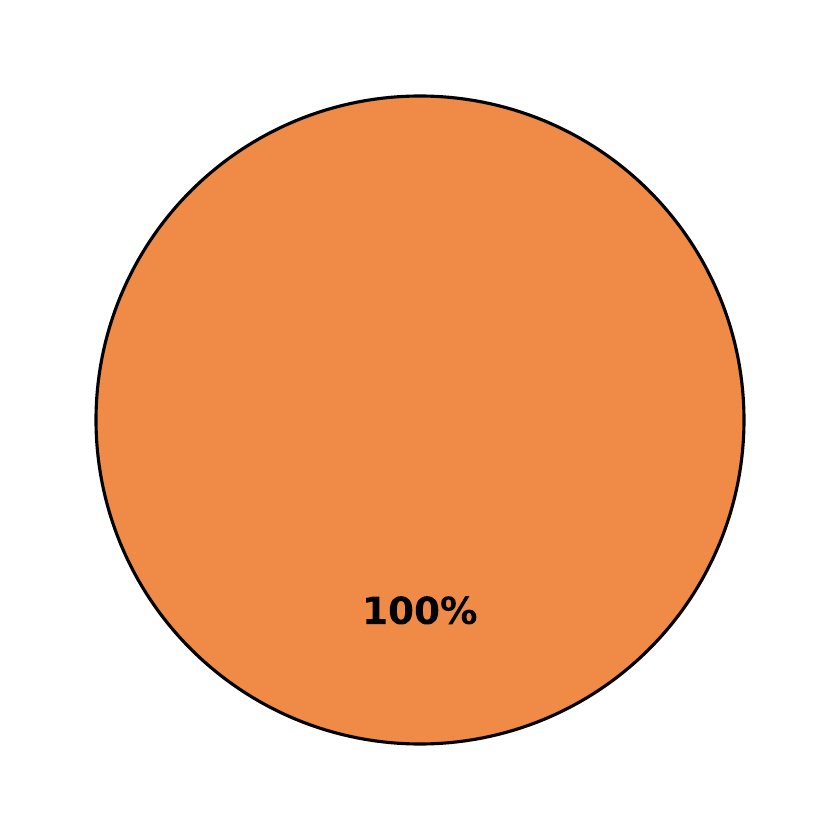}
    \caption{CLOVER}
    \label{subfig:clover_direction}
    \end{subfigure}
    \caption{Proportion of data projections across different components in random directions (LoRA) versus orthogonal directions (PiSSA), as well as all orthogonal directions (CLOVER).}
    \label{fig:lora_pissa_direction}
\end{figure}

1) Without accounting for the scaling effect of singular values, the projection magnitude along the principal singular vector consistently exceeds that in other directions. This observation supports PiSSA's approach, which updates based on the principal singular values and vectors, leading to improved training performance. In contrast, LoRA projects in random directions, resulting in uniform projection magnitudes across all directions.

2) The singular values in the original model reflect the importance of each direction in the pretraining task. The model amplifies the components along directions with larger singular values and suppresses those along smaller singular values. Therefore, it is crucial to consider the scaling effect of singular values. As shown in Figure \ref{subfig:pissa_with_s_direction}, the projection magnitude along the principal singular vector direction increases to 18\%.

3) While more data projections align with the principal singular vector at higher ranks, 82\% of the feature components are still projected onto other directions. In extreme cases, if a task is entirely orthogonal to the vectors used by PiSSA, training on such a task may result in zero gradients, thereby limiting its learning capacity. Under the same rank constraint, 94\% of the feature components in LoRA are projected outside the LoRA adapter, making it more susceptible to the zero-gradient problem. 

Since CLOVER updates across all orthogonal directions, as shown in Figure \ref{subfig:clover_direction} it effectively mitigates this issue. Consequently, CLOVER outperforms both LoRA and PiSSA in multi-task learning, even when using the same or fewer learnable parameters (Section \ref{subsec:clover_for_finetuning_on_commonsense}).

\subsection{Visualizing Rank Updates}
\label{subsec:full_rank_update}
\begin{figure}[htbp]
    \centering
    \begin{subfigure}[b]{0.32\columnwidth}
    \includegraphics[width=\columnwidth]{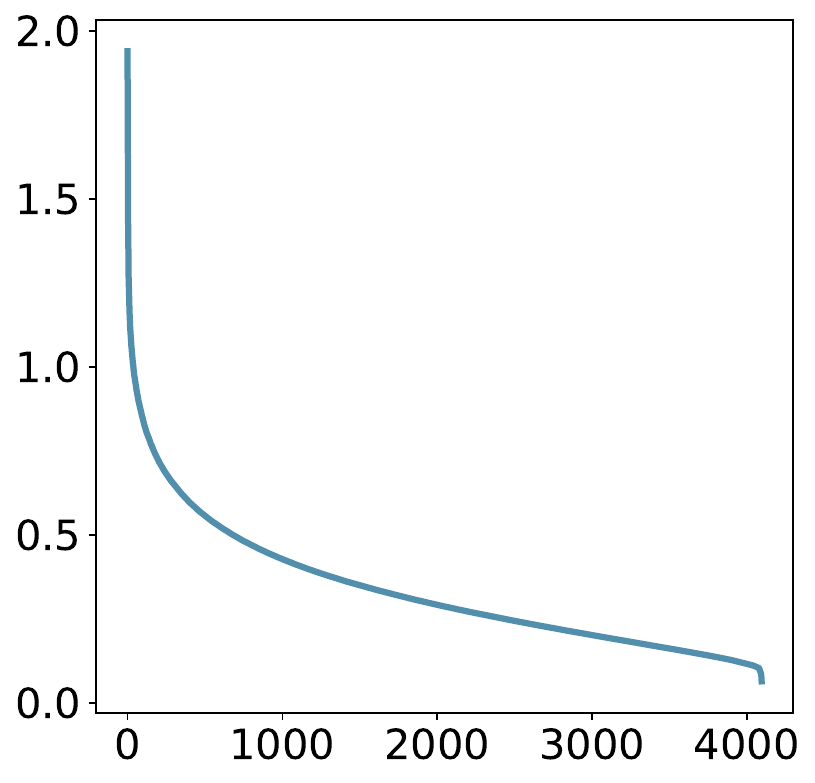}
    \caption{Full Fine-Tuning}
    \label{subfig:fullft_delta_W}
    \end{subfigure}
    \begin{subfigure}[b]{0.305\columnwidth}
    \includegraphics[width=\columnwidth]{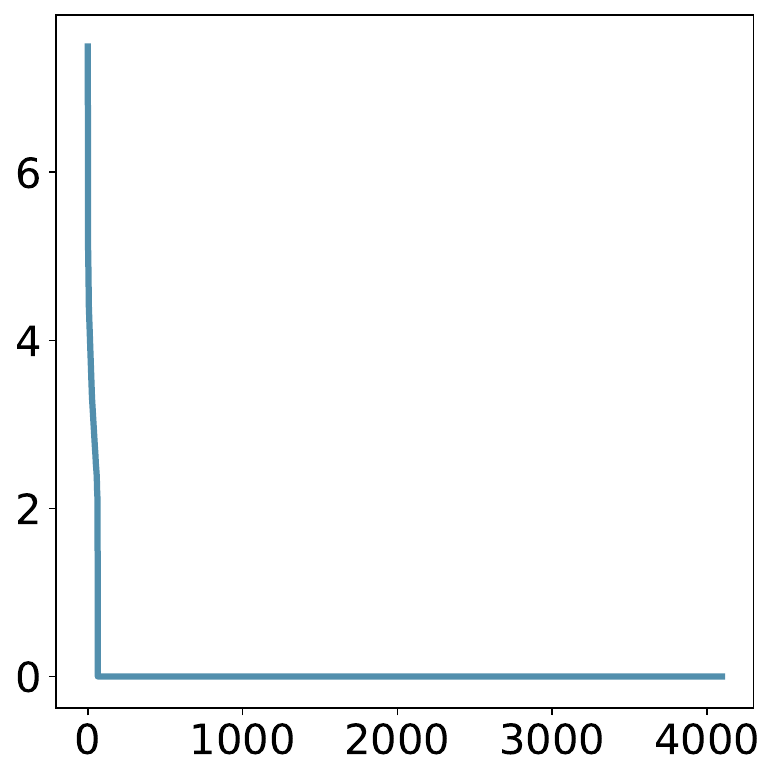}
    \caption{LoRA}
    \label{subfig:pissa_delta_W}
    \end{subfigure}
    \begin{subfigure}[b]{0.32\columnwidth}
    \includegraphics[width=\columnwidth]{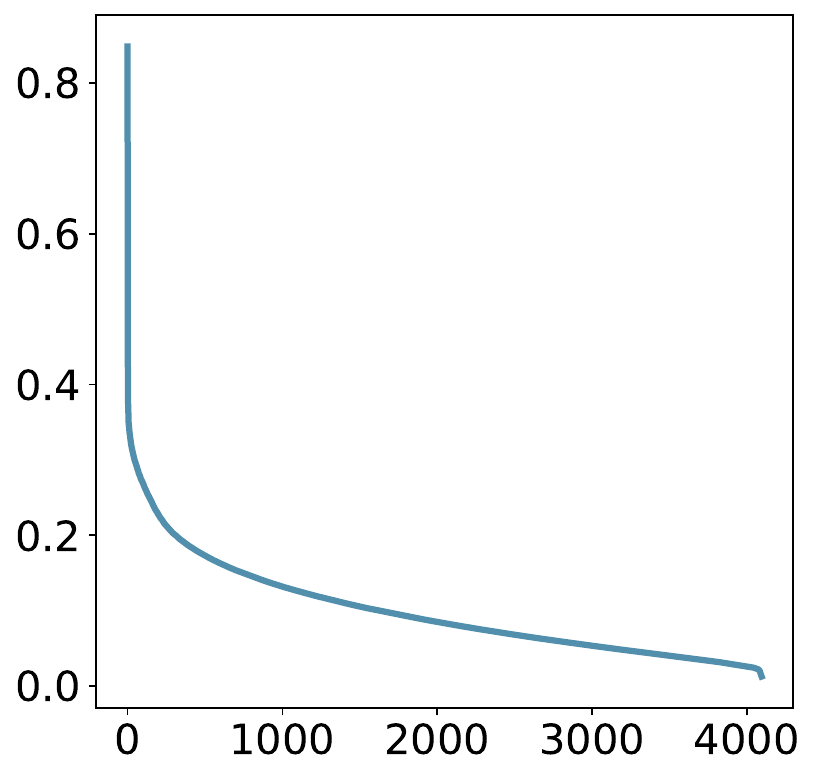}
    \caption{CLOVER}
    \label{subfig:clover_delta_W}
    \end{subfigure}
    \caption{$\Delta W$ is low rank in LoRA, while full rank for Full-Fine-Tuning and CLOVER.}
    \label{fig:low_rank_delta_w}
\end{figure}

To demonstrate CLOVER achieves full-rank updates, we multiply the updated singular values with their corresponding singular vectors and perform SVD on the base model ($S_{QK}$ applied to the Key layer, $S_{VO}$ to the Value layer, and $S_{UD}$ to the Up layer). We take LoRA, and Full Fine-tuning for comparing. Figure \ref{fig:low_rank_delta_w} shows the singular value of the middle layer in LLaMA-2-7B, revealing that CLOVER and Full Fine-tuning achieve full-rank updates, while LoRA is constrained by its low-rank design.

\subsection{CLOVER Avoids Intrusive Dimensions}
\label{subsec:intrusive_dimensions}
\begin{figure}[htbp]
    \centering
    \begin{subfigure}[b]{0.32\columnwidth}
    \includegraphics[width=\columnwidth]{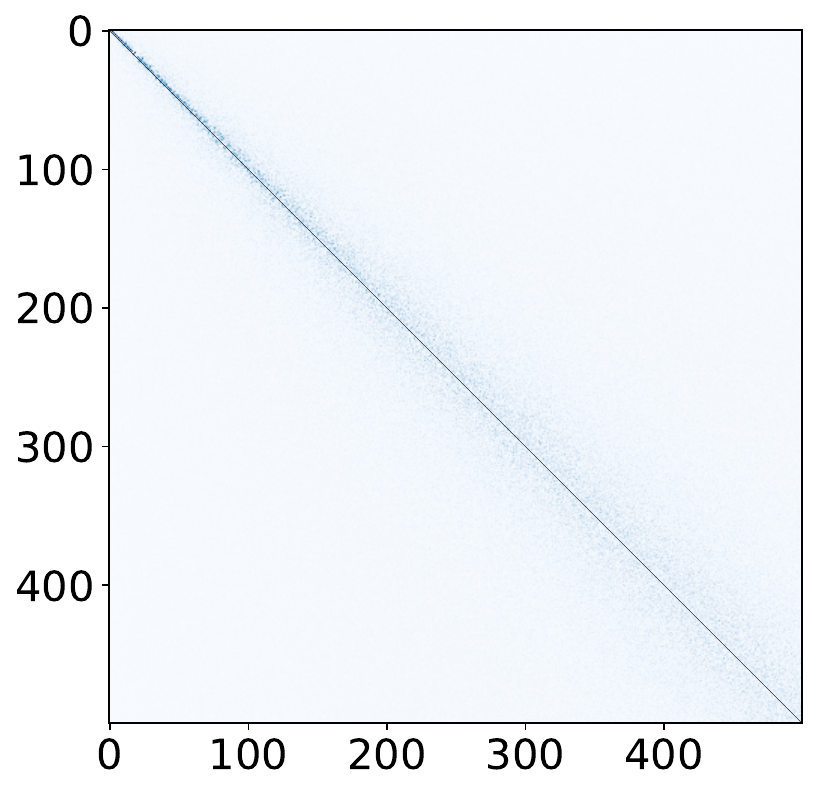}
    \caption{Full Fine-Tuning}
    \label{subfig:fullft_intruder_dimensions}
    \end{subfigure}
    \begin{subfigure}[b]{0.32\columnwidth}
    \includegraphics[width=\columnwidth]{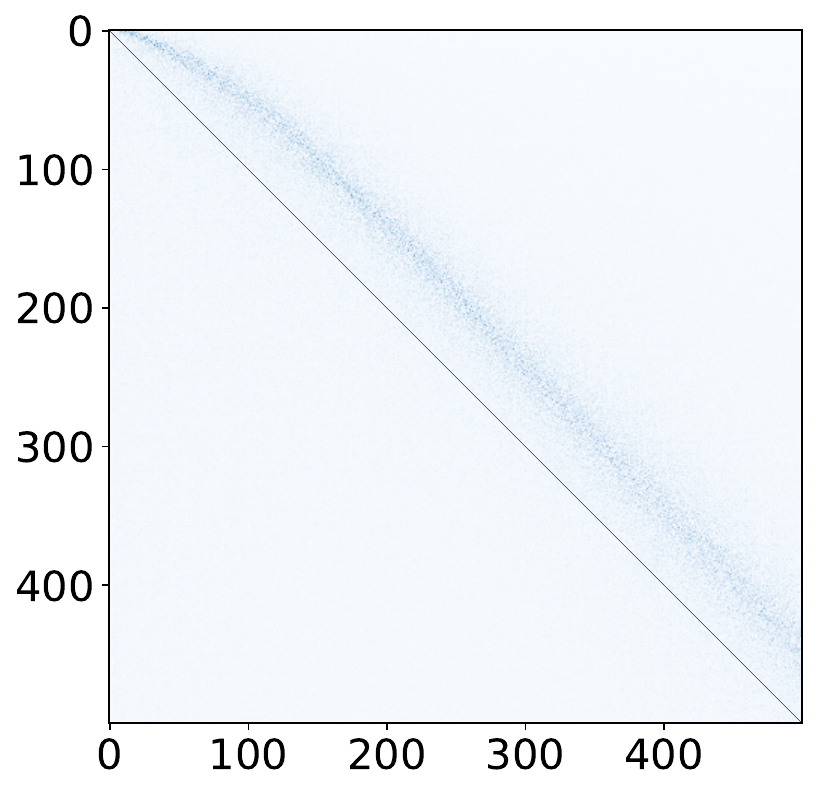}
    \caption{LoRA}
    \label{subfig:lora_intruder_dimensions}
    \end{subfigure}
    \begin{subfigure}[b]{0.32\columnwidth}
    \includegraphics[width=\columnwidth]{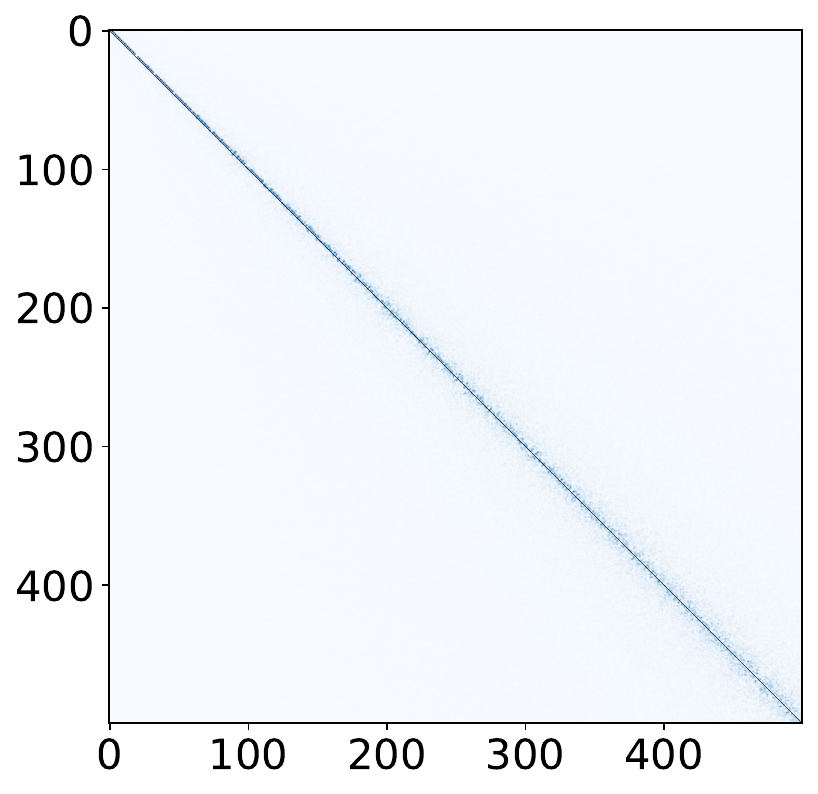}
    \caption{CLOVER}
    \label{subfig:clover_intruder_dimensions}
    \end{subfigure}
    \caption{Intruder dimensions phenomenal in LoRA, which does not exist in Full Fine-Tuning and CLOVER.}
    \label{fig:intruder_dimensions}
\end{figure}

Recent research \cite{shuttleworth2024lora} has highlighted an issue with LoRA, referred to as the “intrusive dimensions” phenomenon. As illustrated in Figure \ref{subfig:lora_intruder_dimensions}, LoRA introduces new random directions into the model, which possess large magnitudes and thus precede all the original singular vectors. The study suggests that these “intrusive dimensions” can degrade the model's performance, exacerbating catastrophic forgetting during continual learning with LoRA. 
In contrast, CLOVER addresses this issue by fixing all orthogonal bases and updating only the vector combinations. As a result, the changes introduced by CLOVER fine-tuning closely resemble those generated by full parameter fine-tuning, as shown in Figure \ref{subfig:fullft_intruder_dimensions} and Figure \ref{subfig:clover_intruder_dimensions}.

\section{Conclusion and Limitations}
In this paper, we introduce Cross-Layer Orthogonal Vectors (CLVOER), a method that orthogonalizes vectors within attention heads without requiring additional transformation matrices. This orthogonalization process condenses effective parameters into fewer vectors, improving the pruning ratio. By fine-tuning the singular values obtained through orthogonalization, CLVOER learns linear combinations of orthogonal bases, enabling full-rank updates.
When applied to prune 50\% of the attention head parameters in GPT-2XL, CLVOER results in a perplexity that is just one-tenth of that achieved by standard pruning methods. For Whisper-Large-v3, CLVOER removes 46.42\% of the parameters without fine-tuning, while preserving model performance. Furthermore, when used for fine-tuning, CLVOER outperforms state-of-the-art methods such as LoRA, DoRA, HiRA, and PiSSA, achieving superior results with equal or fewer trainable parameters. We also demonstrate how CLVOER removes linear redundancy to facilitate pruning and discuss the necessity of fine-tuning across all orthogonal bases. Visual comparisons of models fine-tuned with different methods further illustrate its effectiveness.

Despite its advantages, CLVOER has some limitations. When nonlinear operations are present between Q-K or V-O pairs (such as with the widely-used RoPE \cite{su2024roformer}), cross-layer orthogonalization is not feasible. In these cases, we instead perform head-wise orthogonalization within the Key layer during fine-tuning. Fortunately, CLVOER Fine-Tuning can apply intra-layer attention head orthogonalization, while CLOVER Pruning remains applicable to many popular models, including DeepSeek \cite{dsvii,dsviii}(which uses Decoupled RoPE), ViT and SDXL (which use absolute positional encoding), and BLOOM \cite{le2023bloom} (which employs Alibi relative positional encoding \cite{press2021train}).
Additionally, as a newly proposed method, our current evaluation focuses primarily on basic pruning tasks and does not include comparisons with other state-of-the-art pruning techniques. However, because CLVOER does not alter the model structure and only updates the initialization method, it can be combined with existing pruning methods to further enhance their effectiveness.

As a novel technique, CLVOER holds considerable promise for future applications. For instance, it could be combined with quantization methods to eliminate outliers, guide pruning and fine-tuning based on data feature directions, or even inspire new model architectures.

\newpage
\section*{Impact Statement}

This paper proposes a cross-layer orthogonal initialization method to guide model pruning and efficient fine-tuning, offering valuable insights for the application and development of large models. Both application directions aim to reduce training and inference costs, lower computational overhead, decrease power consumption, and minimize carbon emissions.


\bibliography{example_paper}
\bibliographystyle{icml2025}

\newpage
\appendix
\onecolumn

\section{Appendix}
\subsection{Cross Layer Orthogonal Vectors in Value and Output layers}
\label{sec:vo_ortho}
In the main text, we only presented the orthogonalization process for the Q-K pair. Here, we provide the method for orthogonalizing the V-O pair.
Additionally, for up-down layers, the output dimension of the Up layer can be reshaped into block number × block size, followed by performing orthogonal decomposition within each block.

\begin{align}
Y &=\text{attn(Q, K)}VW_O, \quad \text{$V=XW_V\in \mathbb{R}^{b\times h\times n\times d}$}\\
& =\text{attn(Q, K)}XW_VW_O, \quad \text{$W_VW_O=W_{VO}=USV\in \mathbb{R}^{h\times D\times D}$}\\
&=\text{attn(Q, K)}XUSV, \quad \text{$S_{[:,r_{vo}:,r_{vo}:]}=S_{VO}\in \mathbb{R}^{h\times r_{vo} \times r_{vo}}=0, r_{vo}\leq d$.}\\
&=\text{attn(Q, K)}XU_{VO}S_{VO}V_{VO}, \quad \text{$U_{VO}\in \mathbb{R}^{D\times h\times r_{vo}}$, $V_{VO}\in \mathbb{R}^{h\times r_{vo} \times D}.$}
\end{align}
Through this series of transformations,  $W_V$  and  $W_O$  can be equivalently replaced by orthogonal vectors  $U_{VO}$  and  $V_{VO}$, along with the diagonal matrix $S_{VO}$.
Since $r_{vo} \leq d$, the singular zero values and their corresponding singular vectors can be safely pruned. After guided pruning, $S_{VO}$ can be merged into $U_{VO}$ and $V_{VO}$, resulting in no additional computational overhead.

\subsection{Hyperparameters}
\label{sec:Hyperparameters}
Table \ref{tab:hypter-parameter} presents a comparison of hyperparameters for different fine-tuning methods on commonsense tasks. The target model remains the same for LoRA, DoRA, HiRA, and PiSSA. However, DoRA introduces an additional magnitude module, leading to a slightly higher parameter count.
In a single layer of LoRA, the trainable parameters are as follows:  

In LoRA, the trainable parameters are: 
\[
\begin{aligned}
    Q &= 4096 \times 32 + 4096 \times 32 \\
    K &= 4096 \times 32 + 4096 \times 32 \\
    V &= 4096 \times 32 + 4096 \times 32 \\
    \text{Up} &= 4096 \times 32 + 11008 \times 32 \\
    \text{Down} &= 4096 \times 32 + 11008 \times 32 
\end{aligned}
\]

The total sum is 1,753,088.  

In CLOVER, the trainable parameters are:  

\[
\begin{aligned}
    QK &= 32 \times 128 \times 128 \\
    VO &= 32 \times 128 \times 128 \\
    UD &= 172 \times 64 \times 64
\end{aligned}
\]

The total sum is also 1,753,088.  

Since CLOVER inserts trainable parameters across layers, we use the Q-K pair notation to represent its target model. When CLVOER updates parameters within an attention head, the number of trainable parameters matches exactly that of LoRA at rank 32. To adjust the number of learnable parameters, CLOVER can either span multiple heads or split a single head into multiple blocks.
Both PiSSA and CLOVER exhibit stable training performance. Therefore, instead of validating every 80 steps, we omit frequent validation, improving training efficiency.
\begin{table}[ht]
    \centering
    \caption{Detailed Training Hyperparameters. Q-K,V-O, U-D means CLVOER update pair of orthogonal vectors.}
    \begin{tabular}{ccccccccc}
    \toprule
    \textbf{Method}& \textbf{Target}  &\makecell{\textbf{Evaluation}\\\textbf{steps}} & \textbf{LR} & \textbf{Scheduler} & \makecell{\textbf{Batch}\\\textbf{size}}  & \makecell{\textbf{Warmup}\\\textbf{Steps}}  & \textbf{Epochs}\\
    \midrule
    LoRA  & Q,K,V,U,D  & 80   & 3e-4 & Linear & 16 & 100 & 3 \\
    DoRA  & Q,K,V,U,D  & 80  & 2e-4 & Linear & 16 & 100 & 3 \\
    HiRA  & Q,K,V,U,D  & 80   & 2e-4/2e-4 & Linear & 32 & 100 & 3 \\
    PiSSA & Q,K,V,U,D  & --   & 2e-5 & Linear & 16 & 100 & 3 \\
    CLOVER & Q-K,V-O, U-D &-- & 1e-4 & Linear & 16 & 100 & 3 \\
    \bottomrule
    \end{tabular}
    \label{tab:hypter-parameter}
\end{table}

\subsection{Detail Information of Dataset}
The commonsense reasoning tasks consist of 8 subtasks, each with predefined training and testing sets, as described by LLM-Adapters (Hu et al., 2023). The following table lists the details of each sub-dataset.
\begin{table}[ht]
    \centering
    \caption{Details of datasets for commonsense reasoning tasks.}
    \resizebox{0.99\textwidth}{!}{
    \begin{tabular}{lccl}
    \toprule
    Dataset &  Train & Test & About \\
    \midrule
     BoolQ~\cite{clark2019boolq} &  9,427 &  3,270 &  Naturally occurring yes/no questions from unconstrained settings. \\
     PIQA~\cite{bisk2020piqa} &  16,113 &  1,838 &  Questions with two solutions requiring physical commonsense. \\
     SIQA~\cite{sap2019socialiqa} &  33,410 &  1,954 &  Reasoning about actions and social implications. \\
     HellaSwag~\cite{zellers2019hellaswag} & 39,905 &  10,042 &  Commonsense NLI questions with context and endings. \\ 
     WinoGrande~\cite{sakaguchi2021winogrande} &  \textbf{40,398} &  1,267 &  Fill-in-the-blank task with binary options. \\
     ARC-e~\cite{clark2018think} &  \textbf{2,251} &  2,376 &  Grade-school multiple-choice science questions in Easy sets. \\
     ARC-c~\cite{clark2018think} &  \textbf{1,119} &  1,172 &  Grade-school multiple-choice science questions in Challenge sets. \\
    OBQA~\cite{mihaylov2018can} &  4,957 &  500 &  Questions requiring multi-step reasoning and commonsense knowledge. \\
  \bottomrule
\end{tabular}}
\label{tab:datasets_detail}
\end{table}

For WinoGrande, the original dataset includes multiple partitions: [xs, s, m, l, xl, debiased]. While LLM-Adapters simply concatenated all these partitions, note that the “xl” partition actually includes all others, leading to extensive data duplication. After removing duplicates, the training data is reduced from 63.2K to 40.4K instances.

Additionally, in the LLM-Adapters paper, the training set sizes of ARC\_Challenge and ARC\_Easy were reversed by mistake; here, we correct that error.

\subsection{LibriSpeech Long dataset target transcript}
Below is the reference text of the LibriSpeech Long dataset for comparison.

\textit{
Mr. Quilter is the apostle of the middle classes, and we are glad to welcome his gospel. 
Nor is Mr. Quilter's manner less interesting than his matter. 
He tells us that at this festive season of the year, 
with Christmas and roast beef looming before us, 
similes drawn from eating and its results occur most readily to the mind. 
He has grave doubts whether Sir Frederick Layton's work is really Greek after all, 
and can discover in it but little of rocky Ithaca. 
Linnell's pictures are a sort of Up Guards and Adam paintings, 
and Mason's exquisite idles are as national as a jingo poem. 
Mr. Birkett Foster's landscapes smile at one much 
in the same way that Mr. Carker used to flash his teeth, 
and Mr. John Collier gives his sitter a cheerful slap on the 
back before he says, like a shampooer in a Turkish bath, next man.
}\label{target_audio_translate}

In fact, with Vanilla Pruning ratios of just 22.31\% and 6.69\% for  $W_Q$-$W_K$  and  $W_V$-$W_O$, respectively, the model’s output is already significantly degraded.

\textit{
Mr. Colter is the personal of the classes, 
and we are glad to welcome his gospel. 
Nor is Mr. Colter's manner less interesting than his manner. 
He tells us that at this festive season of the year, 
with Christmas and roast beef looming before us, 
similarly he is drawn from eating and its results occur most readily to the mind. 
He is very dull, so very frequently, and is very Greek after all, 
and can discover in it but little of Rocky Ithaca. 
The Nell's pictures are sort of up-guard to Adam's paintings, 
and Mason's exquisite idylls are as national as a jingle poem. 
Mr. Burke and Foster's landscapes smile at one much 
in the same way as Mr. Parker, Mr. Flash is tits. 
And Mr. John Collier gives his sitter a cheerful slap on the 
back before he says like a shampoo and a Turkish bath, Next man.
}

\subsection{Visualizing more attention heads}
\begin{figure}[ht]
    \centering
    \includegraphics[width=\linewidth]{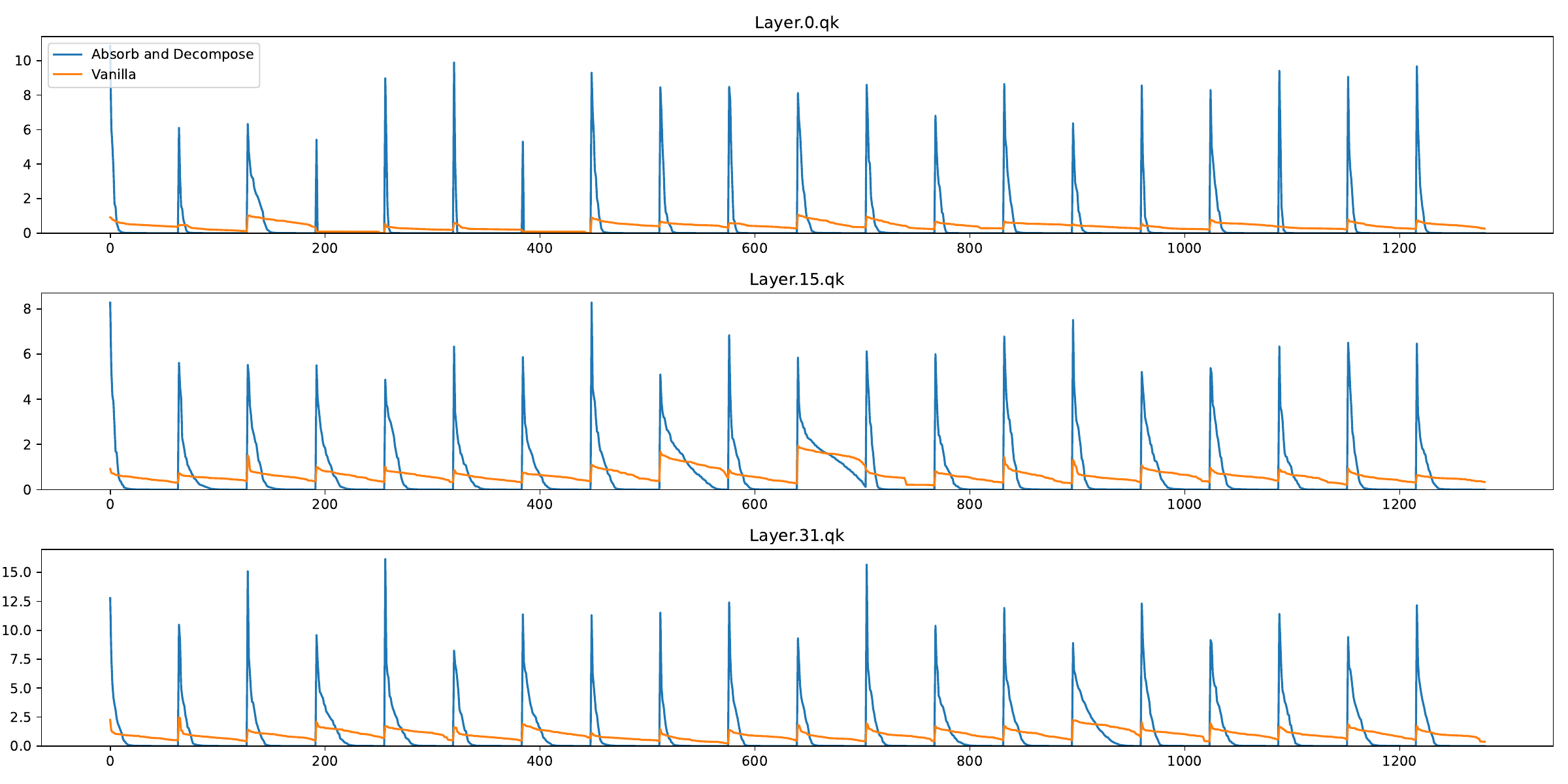}
    \caption{The $L_2$-norm for the 0-th, 15-th, and 31-st attention layers in the Whisper-large-v3 encoder. The blue line represents the results after redundancy removal using the CLOVER method, while the orange line depicts the $L_2$-norm directly computed for each dimension.}
    \label{fig:whisper_qkv}
\end{figure}

\begin{figure}[ht]
    \centering
    \includegraphics[width=\linewidth]{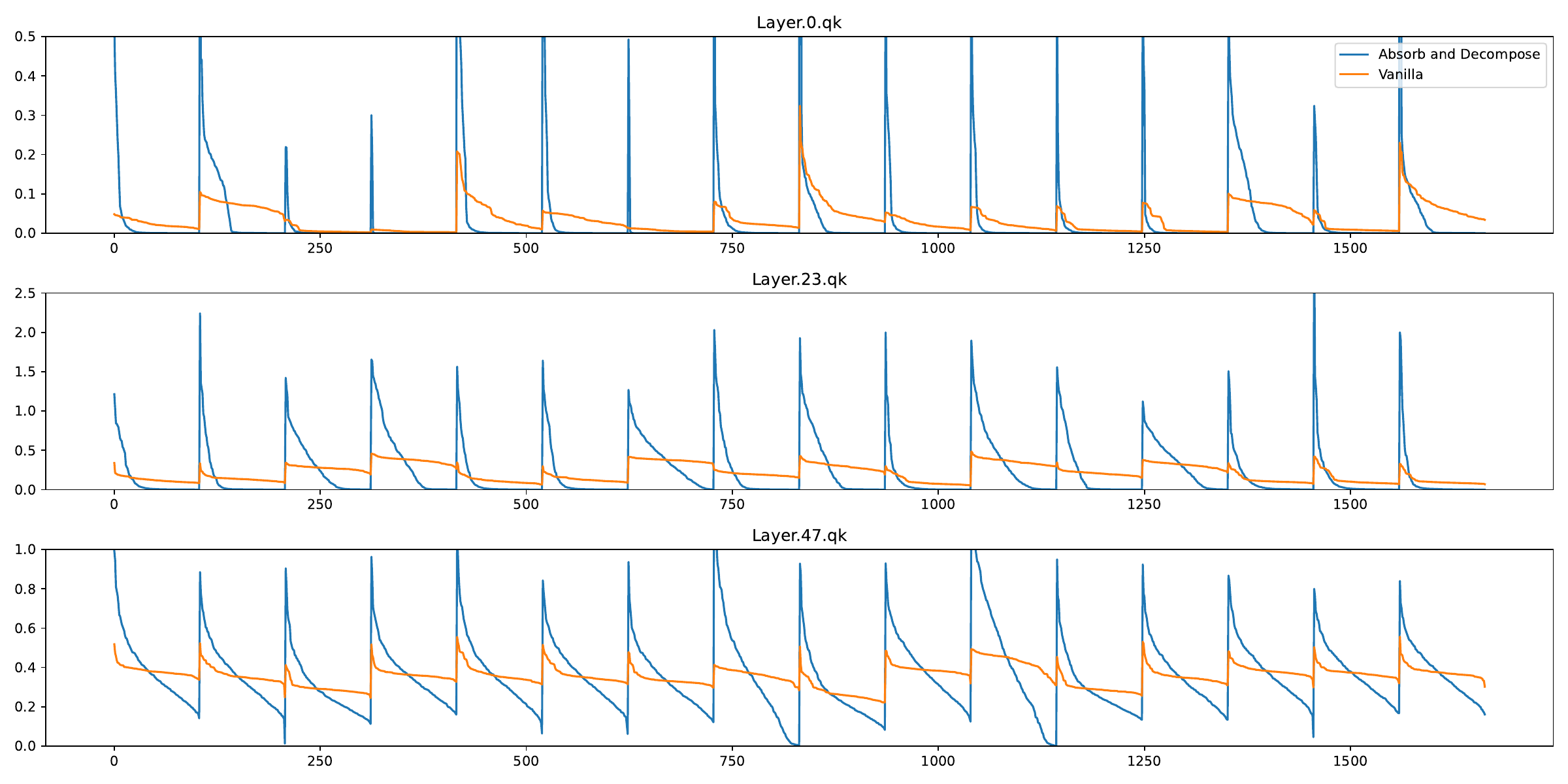}
    \caption{The $L_2$-norm for the 0-th, 15-th, and 31-st attention layers in the ViT-bigG. The blue line represents the results after redundancy removal using the CLOVER method, while the orange line depicts the $L_2$-norm directly computed for each dimension.}
    \label{fig:vit_qkv}
\end{figure}

In Section \ref{subsec:visualize_pruning_models}, we only presented the first attention head in the first layer. Here, we provide a broader view by showcasing more attention heads.
Figure \ref{fig:whisper_qkv} illustrates the $L_2$ norm of all Q-K heads in the first, middle, and last layers of Whisper-Large-v3.
Figure \ref{fig:vit_qkv} shows the $L_2$ norm of all Q-K heads in the first, middle, and last layers of ViT-bigG.

From these figures, we can observe that CLOVER consistently represents the entire attention head with fewer orthogonal bases across all layers and all attention heads. This property forms the foundation of CLVOER’s effectiveness in enhancing pruning.

\end{document}